\documentclass[10pt,journal,cspaper,compsoc]{IEEEtran}

\ifCLASSOPTIONcompsoc
  \usepackage[nocompress]{cite}
\else
  \usepackage{cite}
\fi

\usepackage[font=footnotesize,labelfont=sf,textfont=sf]{caption}
\usepackage[footnotesize,sf,SF]{subfigure}
\usepackage{graphicx}
\usepackage{amsmath,amsthm}
\usepackage{amssymb}
\usepackage[mathscr]{euscript}
\usepackage{framed,color}
\usepackage[linesnumbered,ruled]{algorithm2e}
\usepackage{url}
\usepackage{multirow}

\makeatletter
\DeclareRobustCommand\onedot{\futurelet\@let@token\@onedot}
\def\@onedot{\ifx\@let@token.\else.\null\fi\xspace}

\def\eg{\emph{e.g}\onedot} 
\def\ie{\emph{i.e}\onedot} 
 
\def\etc{\emph{etc}\onedot} 
 
\def\etal{\emph{et al}\onedot}

\newtheorem{theorem}{Theorem}
\newtheorem{lemma}{Lemma}
\newtheorem{corollary}[theorem]{Corollary}

\newcommand{\bR}{\mathbf{R}}
\newcommand{\bt}{\mathbf{t}}
\newcommand{\br}{\mathbf{r}}
\newcommand{\bx}{\mathbf{x}}
\newcommand{\by}{\mathbf{y}}
\newcommand{\bgamma}{\boldsymbol{\gamma}}
\newcommand{\rT}{\mathrm{T}}

\DeclareMathOperator*{\argmin}{argmin}

\SetKwBlock{Begin}{loop}{end}
\SetKwBlock{doWhile}{do}{while}

\begin{document}

\title{Go-ICP:  A Globally Optimal Solution to\\3D ICP Point-Set Registration}
\author{
\vspace{10pt}
Jiaolong~Yang,
        Hongdong~Li,
        Dylan~Campbell,
        and~Yunde~Jia
\IEEEcompsocitemizethanks{\IEEEcompsocthanksitem J. Yang is with the Beijing Lab of Intelligent Information Technology, Beijing Institute of Technology (BIT), China, and the Australian National University (ANU), Australia. Email: yangjiaolong@bit.edu.cn.
\IEEEcompsocthanksitem H. Li and D. Campbell are with the Australian National University (ANU), and National ICT Australia (NICTA). \protect\\Email: hongdong.li@anu.edu.au; dylan.campbell@anu.edu.au.
\IEEEcompsocthanksitem Y. Jia is with the Beijing Lab of Intelligent Information Technology, Beijing Institute of Technology (BIT), China. Email: jiayunde@bit.edu.cn.
}
\vspace{5pt}
}

\markboth{ }%
{Shell \MakeLowercase{\textit{YANG et al.}}: Go-ICP:  A Globally Optimal Solution to 3D ICP Point-Set Registration}

\IEEEcompsoctitleabstractindextext{
\begin{abstract}
The Iterative Closest Point (ICP) algorithm is one of the most widely used methods for point-set registration.  However, being based on local iterative optimization, ICP is known to be susceptible to local minima. Its performance critically relies on the quality of the initialization and only local optimality is guaranteed.  This paper presents the first globally optimal algorithm, named Go-ICP, for Euclidean (rigid) registration of two 3D point-sets under the $L_2$ error metric defined in ICP. The Go-ICP method is based on a branch-and-bound (BnB) scheme that searches the entire 3D motion space $SE(3)$.  By exploiting the special structure of $SE(3)$ geometry, we derive novel upper and lower bounds for the registration error function. Local ICP is integrated into the BnB scheme, which speeds up the new method while guaranteeing global optimality.  We also discuss extensions, addressing the issue of outlier robustness. The evaluation demonstrates that the proposed method is able to produce reliable
registration results regardless of the initialization. Go-ICP can be applied in scenarios where an optimal solution is desirable or where a good initialization is not always available.
\end{abstract}
\vspace{5pt}
\vspace{5pt}
}

\maketitle

\IEEEdisplaynotcompsoctitleabstractindextext

\section{Introduction}\label{sec:introduction}
\noindent Point-set registration is a fundamental problem in computer and robot vision. Given two sets of points in different coordinate systems, or equivalently in
the same coordinate system with different poses, the goal is to find the transformation that best aligns one of the point-sets to the other. Point-set registration plays an important role in many vision applications. Given multiple partial scans of an object, it can be applied to merge them into a complete 3D model~\cite{blais1995registering,huber2003fully}. In object recognition, fitness scores of a query object with respect to existing model objects can be measured with registration results~\cite{johnson1999using,belongie2002shape}. In robot navigation, localization can be achieved by registering the current view into the global environment \cite{nuchter20076d,pomerleau2013comparing}. Given cross-modality data acquired from different sensors with complementary information, registration can be used to fuse the data~\cite{makela2002review,zhao2005alignment} or determine the relative poses between these sensors~\cite{yang2013single,geiger2012automatic}.

Among the numerous registration methods proposed in literature, the Iterative Closest Point (ICP) algorithm \cite{besl1992method,yang1991object,zhang1994iterative}, introduced in the early 1990s, is the most well-known algorithm for efficiently registering two 2D or 3D point-sets under Euclidean (rigid) transformation. Its concept is simple and intuitive: given an initial transformation (rotation and translation), it alternates between building closest-point correspondences under the current transformation and estimating the transformation with these correspondences, until convergence. Appealingly, point-to-point ICP is able to work directly on the raw point-sets, regardless of their intrinsic properties (such as distribution, density and noise level). Due to its conceptual simplicity, high usability and good performance in practice, ICP and its variants are very popular and have been successfully applied in numerous real-world tasks (\cite{newcombe2011kinectfusion, seitz2006comparison, makela2002review}, for example).

However, ICP is also known for its susceptibility to the problem of local minima, due to the non-convexity of the problem as well as the local iterative procedure it adopts.  Being an iterative method, it requires a good initialization, without which the algorithm may easily become trapped in a local minimum.  If this occurs, the solution may be far from the true (optimal) solution, resulting in erroneous estimation.  More critically, there is no reliable way to tell whether or not it is trapped in a local minimum.

To deal with the issue of local minima, previous efforts have been devoted to widening the basin of convergence~\cite{fitzgibbon2003robust,tsin2004correlation}, performing heuristic and non-deterministic global search~\cite{sandhu2010point,silva2005precision} and utilizing other methods for coarse initial alignment~\cite{rusu2009fast,makadia2006fully}, \etc. However, global optimality cannot be guaranteed with these approaches. Furthermore, some methods, such as those based on feature matching, are not always reliable or even applicable when the point-sets are not sampled densely from smooth surfaces.

This work is, to the best of our knowledge, the first to propose a globally optimal solution to the Euclidean registration problem defined by ICP in 3D. The proposed method always produces the exact and globally optimal solution, up to the desired accuracy. Our method is named the \emph{Globally Optimal ICP}, abbreviated to \emph{Go\nobreakdash-ICP}.

We base the Go-ICP method on the well-established Branch-and-Bound (BnB) theory for global optimization. Nevertheless, choosing a suitable domain parametrization for building a tree structure in BnB and, more importantly, deriving efficient error bounds based on the parametrization are both non-trivial. Our solution is inspired by the $SO(3)$ space search technique proposed in Hartley and Kahl~\cite{hartley2007global} as well as Li and Hartley~\cite{li20073d}. We extend it to $SE(3)$ space search and derive novel bounds of the 3D registration error. Another feature of the Go-ICP method is that we employ, as a subroutine, the  conventional (local) ICP algorithm within the BnB search procedure.
The algorithmic structure of the proposed method can be summarized as follows.
\vspace{0pt}
\begin{framed}
\vspace{-0pt}
\noindent \emph{Use BnB to search the space of $SE(3)$}

\hangafter 0
\hangindent 1.5em
\noindent \emph{Whenever a better solution is found, call ICP initialized at this solution to refine (reduce) the objective function value.  Use ICP's result as an updated upper bound to continue the BnB.}

\hangindent 0em
\noindent \emph{Until convergence.}
\vspace{-0pt}
\end{framed}
\vspace{0pt}

Our error metric strictly follows that of the original ICP algorithm, that is, minimizing the $L_2$~norm of the closest-point residual vector. We also show how a trimming strategy can be utilized to handle outliers. With small effort, one can also extend the method with robust kernels or robust norms. A preliminary version of this work was presented as a conference paper~\cite{yang2013goicp}.

\subsection{Previous Work}
There is a large volume of work published on ICP and other registration techniques, precluding us from giving a comprehensive list. Therefore, we will focus below on some relevant Euclidean registration works addressing the local minimum issue in 2D or 3D. For other papers, the reader is referred to two surveys on ICP variants~\cite{rusinkiewicz2001efficient,pomerleau2013comparing}, a recent survey on 3D point cloud and mesh registration~\cite{tam2013registration}, an overview of 3D registration~\cite{castellani20123d} and the references therein.

\vspace{4pt}
\noindent\textbf{\emph{{Robustified Local Methods.}}} To improve the robustness of ICP to poor initializations, previous work has attempted to enlarge the basin of convergence by smoothing out the objective function. Fitzgibbon~\cite{fitzgibbon2003robust} proposed the LM-ICP method where the ICP error was optimized with the Levenberg--Marquardt algorithm~\cite{more1978levenberg}. Better convergence than ICP was observed, especially with the use of robust kernels.

It was shown by Jian and Vemuri~\cite{jian2005robust} that if the point-sets are represented with Gaussian Mixture Models (GMMs), ICP is related to minimizing the Kullback-Leibler divergence of two GMMs. Although improved robustness to outliers and poor initializations could be achieved by GMM-based techniques~\cite{jian2005robust,tsin2004correlation,myronenko2010point,campbell2015adaptive}, the optimization was still based on local search. Earlier than these works, Rangarajan \etal~\cite{rangarajan1997robust} presented a SoftAssign algorithm which assigned Gaussian weights to the points and applied deterministic annealing on the Gaussian variance. Granger and Pennec~\cite{granger2002multi} proposed an algorithm named Multi-scale EM-ICP where an annealing scheme on GMM variance was also used. Biber and Stra{\ss}er~\cite{biber2003normal} developed the Normal Distributions Transform (NDT) method, where Gaussian models were defined for uniform cells in a spatial grid. Magnusson \etal~\cite{magnusson2009evaluation} experimentally showed that NDT was more robust to poor initial alignments than ICP.

Some methods extend ICP by robustifying the distance between points.
For example, Sharp \etal~\cite{sharp2002icp} proposed the additional use of invariant feature descriptor distance; Johnson and Kang~\cite{johnson1999registration} exploited color distances to boost the performance.

\vspace{0.06in}
\noindent\textbf{\emph{{Global Methods.}}}
To address the local minima problem, global registration methods have also been investigated. A typical family adopts stochastic optimization such as Genetic Algorithms~\cite{silva2005precision,robertson2002parallel}, Particle Swam Optimization~\cite{wachowiak2004approach}, Particle Filtering~\cite{sandhu2010point}  and Simulated Annealing schemes~\cite{blais1995registering,papazov2011stochastic}. While the local minima issue is effectively alleviated, global optimality cannot be guaranteed and initializations still need to be reasonably good as otherwise the parameter space is too large for the heuristic search.

Another class of global registration methods introduces shape descriptors for coarse alignment. Local descriptors, such as Spin Images~\cite{johnson1999using}, Shape Contexts~\cite{belongie2002shape}, Integral Volume~\cite{gelfand2005robust} and Point Feature Histograms~\cite{rusu2009fast} are invariant under specific transformations. They can be used to build sparse feature correspondences, based on which the best transformation can be found with random sampling~\cite{rusu2009fast}, greedy algorithms~\cite{johnson1999using}, Hough Transforms~\cite{woodford2014demisting} or BnB algorithms~\cite{gelfand2005robust,bazin2012globally}. Global shape descriptors, such as Extended Gaussian Images (EGI)~\cite{makadia2006fully}, can be used to find the best transformation maximizing descriptor correlation. These methods are often robust and can efficiently register surfaces where the descriptor can be readily computed.

Random sampling schemes such as RANSAC~\cite{fischler1981random} can also be used to register raw point clouds directly. Irani and Raghavan~\cite{irani1999combinatorial} randomly sampled 2-point bases to align 2D point-sets using similarity transformations. For 3D, Aiger \etal~\cite{aiger20084} proposed a 4PCS algorithm that sampled coplanar 4-points, since congruent coplanar 4-point sets can be efficiently extracted with affine invariance.

\vspace{0.06in}
\noindent\textbf{\emph{{Globally Optimal Methods.}}}
Registration methods that guarantee optimality have been published in the past, albeit in a smaller number. Most of them are based on BnB algorithms. For example, geometric BnB has been used for 2D image pattern matching \cite{breuel2003implementation,mount1999efficient,pfeuffer2012discrete}. These methods share a similar structure with ours: given each transformation sub-domain, determine for each data point the uncertainty region, based on which the objective function bounds are derived and the BnB search is applied. However, despite uncertainty region computation with various 2D transformations has been extensively explored, extending them to 3D is often impractical due to the heightened complexity~\cite{breuel2003implementation}.

For 3D registration, Li and Hartley~\cite{li20073d} proposed using a Lipschitzized $L_2$ error function that was minimized by BnB. However, this method makes unrealistic assumptions that the two point-sets are of equal size and that the transformation is pure rotation. Olsson \etal~\cite{olsson2009branch} obtained the optimal solution to simultaneous point-to-point, point-to-line and point-to-plane registration using BnB and bilinear relaxation of rotation quaternions. This method, although related to ours, requires known correspondences. Recently, Bustos \etal~\cite{bustos2014fast} proposed searching $SO(3)$ space for optimal 3D geometric matching, assuming known translation. Efficient run-times were achieved using stereographic projection techniques.

Some optimal 3D registration methods assume a small number of putative correspondences, and treat registration as a correspondence outlier removal problem. For example, to minimize the overall pairwise distance error, Gelfand~\etal~\cite{gelfand2005robust} applied BnB to assign one best corresponding model point for each data point. A similar idea using pairwise consistency was proposed by Enqvist \etal~\cite{enqvist2009optimal}, where the inlier-set maximization was formulated as an NP-hard graph vertex cover problem and solved using BnB. Using angular error, Bazin \etal~\cite{bazin2012globally} solved a similar correspondence inlier-set maximization problem via $SO(3)$ space search assuming known translation. Enqvist and Kahl~\cite{enqvist2008robust} optimally solved camera pose in $SE(3)$ via BnB. However, the key insight is that with pre-matched correspondences, their pairwise constraint (also used in \cite{enqvist2009optimal}) enabled a single translation BnB in $\mathbb{R}^3$ to solve the $SE(3)$ problem.

\vspace{0.02in}
\textbf{In this paper}, we optimally solve the 3D Euclidean registration problem with both rotation and translation. The proposed Go-ICP method is able to work directly on raw sparse or dense point-sets (which may be sub-sampled only for reasons of efficiency), without the need for a good initialization or putative correspondences. The method is related to the idea of $SO(3)$ space search, as proposed in \cite{hartley2007global,li20073d} and extended in \cite{ruland2012globally,bazin2012globally,yang2014optimal}, \etc. We extend the 3-dimensional $SO(3)$ search to 6-dimensional $SE(3)$ search, which is much more challenging.

\section{Problem Formulation}\label{sec:formulation}
In this paper we define the $L_2$-norm registration problem in the same way as in the standard point-to-point ICP algorithm. Let two 3D point-sets $\mathcal{X}=\{\mathbf{x}_i\},i=1,...,N$ and $\mathcal{Y}=\{\mathbf{y}_j\},j=1,...,M$, where $\mathbf{x}_i,\mathbf{y}_j\in\mathbb{R}^3$ are point coordinates, be the \emph{data} point-set and the \emph{model} point-set respectively.  The goal is to estimate a rigid motion with rotation $\bR\!\in\!SO(3)$ and translation $\bt\!\in\!\mathbb{R}^3$, which minimizes the following $L_2$-error $E$,
\begin{equation}
    E(\bR,\mathbf{t}) = \sum_{i=1}^N e_i(\bR,\mathbf{t})^2 = \sum_{i=1}^N \|\bR\bx_i+\bt-\by_{j^*}\|^2
    \label{eq:registrationerror}
\end{equation}
where $e_i(\bR,\mathbf{t})$ is the per-point residual error for $\bx_i$. Given $\bR$ and $\mathbf{t}$, the point $\mathbf{y}_{j^*}\in{\mathcal{Y}}$ is denoted as the optimal correspondence of $\bx_i$, which is the closest point to the transformed $\bx_i$ in $\mathcal{Y}$, \ie
\begin{equation}
j^*=\argmin_{j\in\{1,..,M\}}\|\bR\mathbf{x}_i+\mathbf{t}-\mathbf{y}_{j}\|.
\label{eq:closestpoint}
\end{equation} Note the short-hand notation used here: $j^*$ varies as a function of $(\bR,\bt)$ and also depends on $\bx_i$.

Equations (\ref{eq:registrationerror}) and (\ref{eq:closestpoint}) actually form a well-known \emph{chicken-and-egg} problem: if the true correspondences are known \emph{a priori}, the transformation can be optimally solved in closed-form \cite{horn1987closed,arun1987least}; if the optimal transformation is given, correspondences can also be readily found. However, the joint problem cannot be trivially solved. Given an initial transformation $(\bR,\bt)$, ICP iteratively solves the problem by alternating between estimating the transformation with (\ref{eq:registrationerror}), and finding closest-point matches with (\ref{eq:closestpoint}).  Such an iterative scheme guarantees convergence to a local minimum~\cite{besl1992method}.

\begin{figure}[!t]
\begin{center}
\subfigure{
\includegraphics[width=0.33\textwidth]{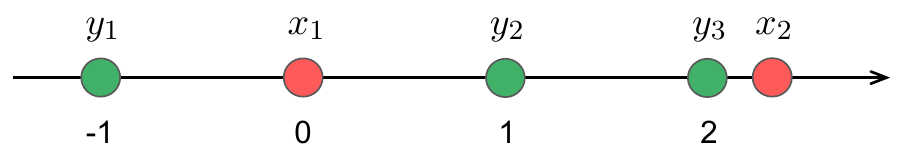}}
\subfigure{
\includegraphics[width=0.23\textwidth]{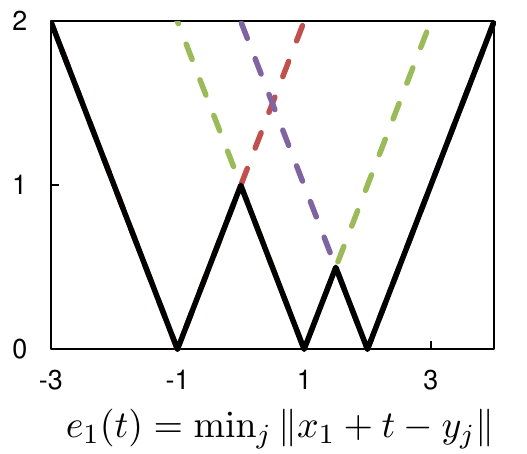}}
\subfigure{
\includegraphics[width=0.23\textwidth]{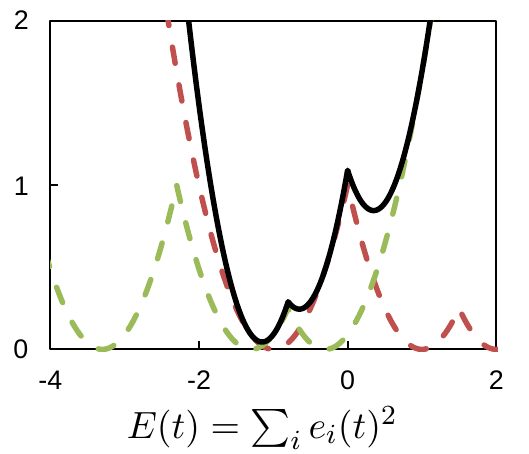}}
\vspace{0pt}
\caption{Nonconvexity of the registration problem. \textbf{Top}: two 1D point-sets $\{x_1,x_2\}$ and $\{y_1,y_2,y_3\}$. \textbf{Bottom-left}: residual error (closest-point distance) for $x_1$ as a function of translation $t$; the three dashed curves are $\|x_1\!+\!t\!-\!y_j\|$ with $j\!=\!1,2,3$ respectively. \textbf{Bottom-right}: the overall $L_2$ registration error; the two dashed curves are $e_i(t)^2$ with $i\!=\!1,2$ respectively. The residual error functions are nonconvex, thus the $L_2$ error function is also nonconvex.}\label{fig:nonconvexity}
\end{center}
\end{figure}

\vspace{0.06in}
\noindent\textbf{\emph{{(Non-)Convexity Analysis.}}}
It is easy to see from (\ref{eq:registrationerror}) that the transformation function denoted by $T_x(p)$ affinely transforms a point $x$ with parameters $p$, thus the residual function $e(p)=d(T_x(p))$ is convex provided that \emph{domain $D_p$ is a convex set (Condition~1)} and $d(x)=\inf_{y\in \mathcal{Y}}\|x-y\|$ is convex. Moreover, it has been shown in \cite{boyd2004convex} and further in \cite{olsson2009branch} that $d(x)$ is convex if and only if \emph{$\mathcal{Y}$ is a convex set (Condition~2)}. For registration with pure translation, Condition~1 can be satisfied as the domain $D_p$ is $\mathbb{R}^3$. However, $\mathcal{Y}$ is often a discrete point-set sampled from complex surfaces and is thus rarely a convex set, violating Condition~2. Therefore, $e(p)$ is nonconvex. Figure~\ref{fig:nonconvexity} shows a 1D example. For registration with rotation, even Condition~1 cannot be fulfilled, as the rotation space induced by the quadratic orthogonality constraints $\bR\bR^\mathrm{T}=\mathbf{I}$ is clearly not a convex set.

\vspace{0.06in}
\noindent\textbf{\emph{Outlier Handling.}}
As is well known, $L_2$-norm least squares fitting is susceptible to outliers.
A small number of outliers may lead to erroneous registration, even if the global optimum is achieved.
There are many strategies to deal with outliers \cite{rusinkiewicz2001efficient,champleboux1992accurate,fitzgibbon2003robust,jian2005robust,chetverikov2005robust}.
In this paper, a trimmed estimator is used to gain outlier robustness similar to \cite{chetverikov2005robust}. To streamline the presentation and mathematical derivation, we defer the discussion to Sec.~\ref{sec:outlier}. For now we assume there are no outliers and focus on minimizing (\ref{eq:registrationerror}).

\section{The Branch and Bound Algorithm}\label{sec:bnb}
The BnB algorithm is a powerful global optimization technique that can be used to solve nonconvex and NP-hard problems~\cite{lawler1966branch}. Although existing BnB methods work successfully for 2D registration, extending them to search $SE(3)$ and solve 3D rigid registration has been much more challenging~\cite{breuel2003implementation,li20073d}. In order to apply BnB to 3D registration, we must consider \emph{i}) how to parametrize and branch the domain of 3D motions (Sec.~\ref{sec:domain}), and \emph{ii}) how to efficiently find upper bounds and lower bounds (Sec.~\ref{sec:bounds}).

\subsection{Domain Parametrization}\label{sec:domain}
Recall that our goal is to minimize the error $E$ in (\ref{eq:registrationerror}) over the domain of all feasible 3D motions (the $SE(3)$ group, defined by $SE(3)=SO(3)\times\mathbb{R}^3$). Each member of $SE(3)$ can be minimally parameterized by 6 parameters (3 for rotation and 3 for translation).

Using the \emph{angle-axis representation}, each rotation can be represented as a 3D vector $\br$, with axis $\br/\|\br\|$ and angle $\|\br\|$. We use $\bR_\br$ to denote the corresponding rotation matrix for $\br$. The 3x3 matrix $\bR_\br\in SO(3)$ can be obtained by the matrix exponential map as
\begin{equation}\label{eq:rodrigues}
\bR_\br = \exp([\,\br\,]_{\times}) = \mathbf{I}\!+\!\frac{[\,\br\,]_{\times} \!\sin{\|\br\|}}{\|\br\|} \!+\! \frac{[\,\br\,]^2_{\times}(1\!-\!\cos{\|\br\|})}{\|\br\|^2}
\end{equation}
\noindent where $[\,\cdot\,]_{\times}$ denotes the skew-symmetric matrix representation
\begin{equation}
[\,\br\,]_{\times} =
\left[\!\begin{array}{ccc}
0 & -r^3 & r^2 \\
r^3 & 0 & -r^1 \\
-r^2 & r^1 & 0
\end{array}\!\right]
\end{equation}
\noindent where $r^i$ is the $i$th element in $\br$. Equation~\ref{eq:rodrigues} is also known as the \emph{Rodrigues' rotation formula}~\cite{hartley2004}. The inverse map is given by the matrix logarithm as
\begin{equation}\label{eq:}
[\,\br\,]_{\times} = \log{\bR_\br}=\frac{\|\br\|}{2\sin{\|\br\|}}(\bR_\br-\bR_{\br}^{\rT})
\end{equation}
\noindent where $\|\br\|\!=\!\arccos{\big((\mathrm{trace}(\bR_\br)\!-\!1)/2\big)}$.
With the angle-axis representation, the entire 3D rotation space can be compactly represented as a solid radius-$\pi$ ball in $\mathbb{R}^3$. Rotations with angles less than (or, equal to) $\pi$ have unique (or, two) corresponding angle-axis representations on the interior (or, surface) of the ball. For ease of manipulation, we use the minimum cube $[-\pi,\pi]^3$ that encloses the $\pi$-ball as the rotation domain.

For the translation part, we assume that the optimal translation lies within a bounded cube $[-\xi,\xi]^3$, which may be readily set by choosing a large number for $\xi$.

\begin{figure}[!t]
\begin{center}
\subfigure[Rotation domain]{
\includegraphics[width=0.16\textwidth]{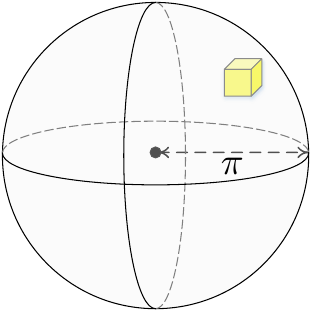}}\ \ \ \ \ \ \ \
\subfigure[Translation domain]{
\includegraphics[width=0.16\textwidth]{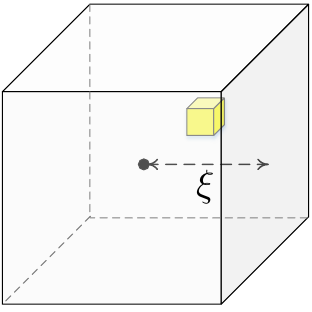}}
\vspace{-2pt}
\caption{$SE(3)$ space parameterization for BnB. \textbf{Left}: the rotation space $SO(3)$ is parameterized in a solid radius-$\pi$ ball with the angle-axis representation. \textbf{Right}: the translation is assumed to be within a 3D cube $[-\xi,\xi]^3$ where $\xi$ can be readily set. The octree data-structure is used to divide (branch) the domains and the yellow box in each diagram represents a sub-cube.}\label{fig:domain}
\end{center}
\vspace{-6pt}
\end{figure}

During BnB search, initial cubes will be subdivided into smaller sub-cubes $C_r$, $C_t$ using the \emph{octree data-structure} and the process is repeated. Figure~\ref{fig:domain} illustrates our domain parametrization.

\section{Bounding Function Derivation}\label{sec:bounds}
For our 3D registration problem, we need to find the bounds of the $L_2$-norm error function used in ICP within a domain $C_r\times C_t$. Next, we will introduce the concept of an \emph{uncertainty radius} as a mathematical preparation, then derive our bounds based on it.

\subsection{Uncertainty Radius}\label{sec:uncertainty}
Intuitively, we want to examine the uncertainty region of a 3D point $\bx$ perturbed by an arbitrary rotation $\br\in C_r$ or a translation $\mathbf{t}\in C_t$. We aim to find a ball, characterised by an uncertainty radius, that encloses such an uncertainty region. We will use the first two lemmas of \cite{hartley2009global} in the following derivation. For convenience, we summarize both lemmas in a single Lemma shown below.\begin{lemma}
    For any vector $\bx$ and two rotations $\bR_{\br}$ and $\bR_{\mathbf{r}_0}$ with $\br$ and $\mathbf{r}_0$ as their angle-axis representations, we have
    \begin{equation}\label{eq:rotationinequality}
    \angle(\bR_{\br}\bx, \bR_{\br_0} \bx)\leqslant\angle(\bR_{\br}, \bR_{\br_0})\leqslant\|\br-\br_0\|,
    \end{equation}
    where $\angle(\bR_{\br},\bR_{\br_0}) = \arccos{\big((\mathrm{trace}(\bR_{\br}^\mathrm{T}\bR_{\br_0})\!-\!1)/2\big)}$ is the angular distance between rotations.
\end{lemma}

The second inequality in (\ref{eq:rotationinequality}) means that the angular distance between two rotations on the $SO(3)$ manifold is less than the Euclidean vector distance of their angle-axis representations in $\mathbb{R}^3$.  Based on this Lemma, uncertainty radii are given as follows.

\begin{figure}[!t]
\begin{center}
\includegraphics[width=0.33\textwidth]{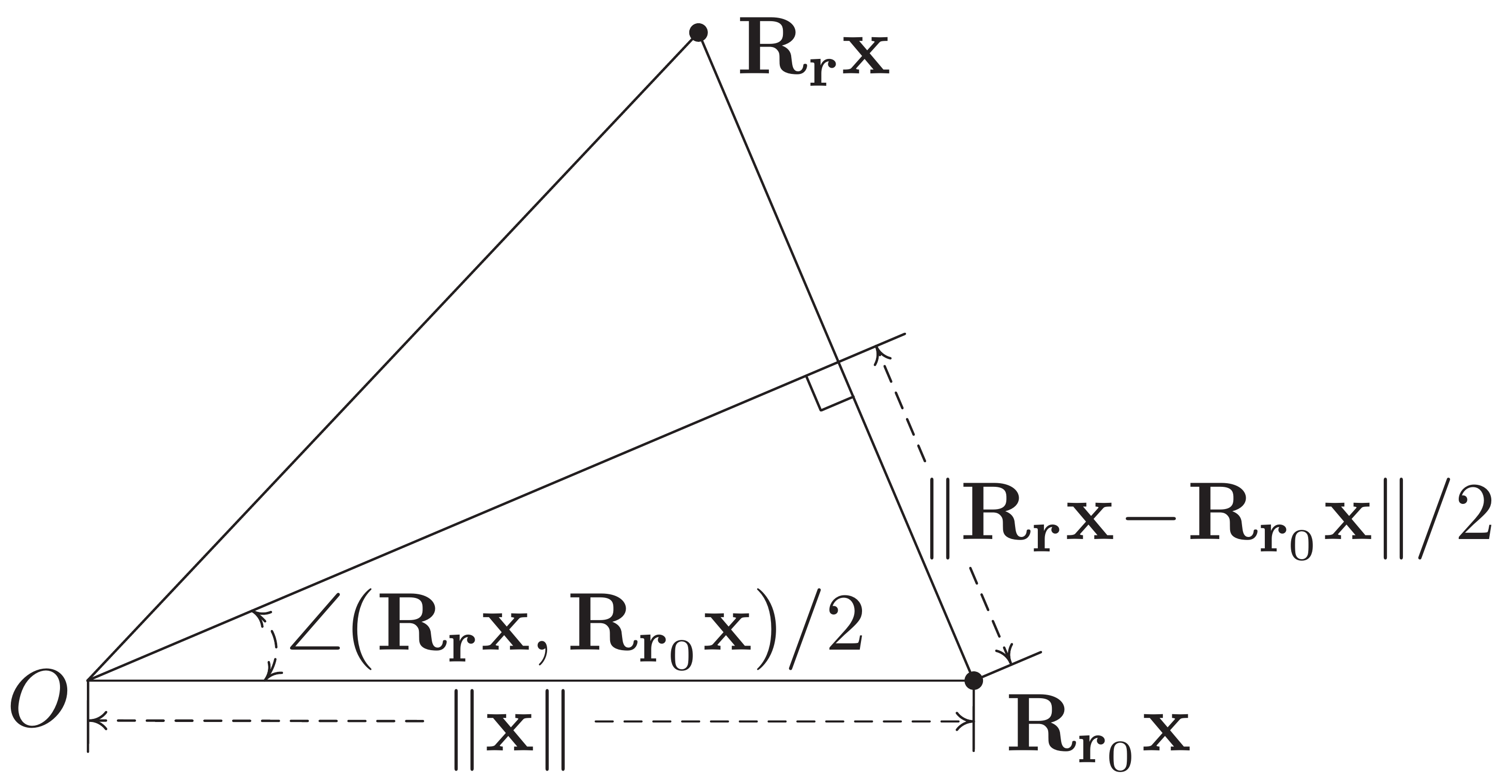}
\caption{Distance computation from $\bR_\br \bx$ to $\bR_{\br_0}\bx$ used in the derivation of the rotation uncertainty radius.}\label{fig:result1eq1}
\end{center}
\end{figure}

\begin{theorem}\label{rs:u_r}(Uncertainty radius)
Given a 3D point $\bx$, a rotation cube $C_r$ of half side-length $\sigma_r$ with $\br_0$ as the center and examining the maximum distance from $\bR_\br \bx$ to $\bR_{\br_0} \bx$, we have $\forall \br\in C_r$,
\begin{equation}\label{eq:rotuncertainty}
\|\bR_\br\bx- \bR_{\br_0} \bx\|
\!\leqslant\! 2\sin(\min({\sqrt{3}\sigma_r}/{2},{\pi}/{2}))\|\bx\|
\!\doteq\! \gamma_r.
\end{equation}
Similarly, given a translation cube $C_t$ with half side-length $\sigma_t$ centered at $\bt_0$, we have $\forall \bt\in C_t$,
\begin{equation}\label{eq:transuncertainty}
\textcolor[rgb]{0.00,0.00,0.00}{\|(\bx+\bt)-(\bx+\bt_0)\|\leqslant \sqrt{3}\sigma_t \doteq \gamma_t.}
\end{equation}
\end{theorem}

\noindent \emph{Proof:} Inequality (\ref{eq:rotuncertainty}) can be derived from
{\begin{align}
\ \ \ \ \ \ \ \ \ \ \ & \|\bR_\br\bx-\bR_{\br_0}\bx\| &\\
& = 2\sin({\angle(\bR_\br \bx, \bR_{\br_0} \bx)}/{2})\|\bx\| & \label{eq:result1eq1}\\
& \leqslant 2\sin(\min({\angle(\bR_\br, \bR_{\br_0})}/{2},{\pi}/{2})) \|\bx\| & \label{eq:result1ieq1}\\
& \leqslant 2\sin(\min({\|\br-\br_0\|}/{2},{\pi}/{2})) \|\bx\| & \label{eq:result1ieq2}\\
& \leqslant 2\sin(\min({\sqrt{3}\sigma_r}/{2},{\pi}/{2}))\|\bx\| \label{eq:result1ieq3}& \end{align}}where (\ref{eq:result1eq1}) is illustrated in Fig.~\ref{fig:result1eq1}. Inequalities (\ref{eq:result1ieq1}), (\ref{eq:result1ieq2}) are based on Lemma 1, and (\ref{eq:result1ieq3}) is from the fact that $\br$ resides in the cube.

Inequality (\ref{eq:transuncertainty}) can be trivially derived via \mbox{$\|(\bx+\bt)-(\bx+\bt_0)\|=\|\bt-\bt_0\|\leqslant \sqrt{3}\sigma_t$}.\qed
\vspace{3pt}

\begin{figure}[!t]
\begin{center}
\subfigure[Rotation uncertainty radius]{
\includegraphics[width=0.23\textwidth]{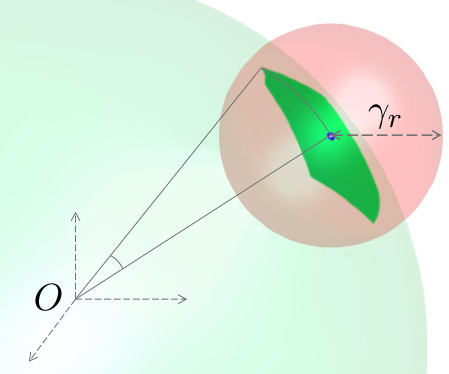}\label{fig:uncertaintyr}}
\subfigure[Translation uncertainty radius]{
\includegraphics[width=0.23\textwidth]{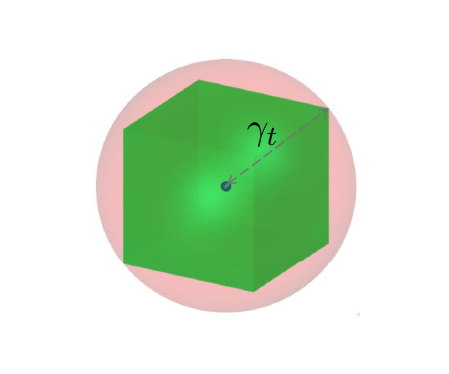}\label{fig:uncertaintyt}}
\vspace{-2pt}
\caption{Uncertainty radii at a point.  \textbf{Left:} rotation uncertainty ball for $C_r$ (in red) with center $\bR_{\br_0}\bx$ (blue dot) and radius $\gamma_r$. \textbf{Right:} translation uncertainty ball for $C_t$ (in red) with center $\bx+\bt_0$ (blue dot) and radius $\gamma_t$.  In both diagrams, the uncertainty balls enclose the range of $\bR_\br\bx$ or $\bx+\bt$ (in green).
\label{fig:uncertainty}}
\end{center}
\end{figure}

We call $\gamma_r$ the rotation uncertainty radius, and $\gamma_t$ the translation uncertainty radius. They are depicted in Fig.~\ref{fig:uncertainty}. Note that $\gamma_r$ is point-dependent, thus we use ${\gamma_r}_i$ to denote the rotation uncertainty radius at $\bx_i$ and the vector $\bgamma_r$ to represent all ${\gamma_r}_i$.
Based on the uncertainty radii, the bounding functions are derived in the following section.

\subsection{Bounding the \boldmath${L_2}$\unboldmath\ Error}\label{sec:errorbound}
Given a rotation cube $C_r$ centered at $\br_0$ and a translation cube $C_t$ centered at $\bt_0$, we will first derive valid bounds of the residual $e_i(\bR,\bt)$ for a single point $\bx_i$.

The upper bound of $e_i$ can be easily chosen by evaluating the error at any $(\br,\bt)\in C_r\times C_t$. Finding a suitable lower bound for the $L_2$ error is a harder task. From Sec.~\ref{sec:uncertainty} we know that, with rotation $\br\in C_r$ (or, translation $\mathbf{t}\in C_t$), a transformed point $\mathbf{x}_i$ will lie in the uncertainty ball centered at $\bR_{\br_0}\bx_i$ (or, $\bx_i+\bt_0$) with radius ${\gamma_r}_i$ (or, $\gamma_t$). For both rotation and translation, it therefore lies in the uncertainty ball centered at $\bR_{\br_0}\bx_i+\bt_0$ with radius ${\gamma_r}_i+\gamma_t$. Now we need to consider the smallest residual error that is possible for $\mathbf{x}_i$. We have the following theorem, which is the cornerstone of the proposed method.

\begin{theorem}(Bounds of per-point residuals) For a 3D motion domain $C_r\times C_t$ centered at $(\br_0,\bt_0)$ with uncertainty radii ${\gamma_r}_i$ and $\gamma_t$, the upper bound $\overline{e_i}$ and the lower bound $\underline{e_i}$ of the optimal registration error $e_i(\bR_\br,\bt)$ at $\bx_i$ can be chosen as
\begin{eqnarray}
&&\!\!\!\!\!\!\!\!\!\overline{e_i} \doteq e_i(\bR_{\br_0},\bt_0), \label{eq:perpointupperbound}\\
&&\!\!\!\!\!\!\!\!\!\underline{e_i} \doteq \max\big(e_i(\bR_{\br_0}, \bt_0)-({\gamma_r}_i\!+\!\gamma_t),0\big). \label{eq:perpointlowerbound}
\end{eqnarray}
\end{theorem}

\noindent\emph{Proof:}
The validity of $\overline{e_i}$ is obvious: error $e_i$ at the specific point $(\br_0,\bt_0)$ must be larger than the minimal error within the domain, \ie $e_i(\bR_{\br_0},\bt_0)\geqslant \min_{\forall(\br,\bt)\in(C_r\times C_t)}e_i(\bR_\br,\bt)$. We now focus on proving the correctness of $\underline{e_i}$.

As defined in (\ref{eq:closestpoint}), the model point $\by_{j^*}\in\mathcal{Y}$ is closest to $(\bR_\br \bx_i+\bt)$.  Let $\by_{j^*_0}$ be the closest model point to $\bR_{\br_0} \bx_i+\bt_0$.  Observe that, $\forall(\br,\bt)\in(C_r\times C_t)$,
\begin{eqnarray}
&&\!\!\!\!\!\!\!\!\!\!\!\!\!\!\!\!\!\!e_i(\bR_\br, \bt) \nonumber\\
&&\!\!\!\!\!\!\!\!\!\!\!\!\!\!\!\!\!\!=\!\|\bR_\br \bx_i\!+\!\bt\!-\!\by_{j^*}\| \\
&&\!\!\!\!\!\!\!\!\!\!\!\!\!\!\!\!\!\!=\!\|( \bR_{\br_0} \bx_i\!+\!\bt_0\!-\!\by_{j^*}\!)\!+(\! \bR_\br \bx_i\!-\! \bR_{\br_0} \bx_i)\!+\!(\bt \!-\!\bt_0)\| \label{eq:eq_auxiliary}\\
&&\!\!\!\!\!\!\!\!\!\!\!\!\!\!\!\!\!\!\geqslant\!\|\bR_{\br_0}\bx_i\!+\!\bt_0\!-\!\by_{j^*}\!\|\!-\!(\|\bR_\br \bx_i\!-\!\bR_{\br_0}\bx_i\|\!+\!\|\bt\!-\!\bt_0\|) \label{eq:ineq_triangle}\\
&&\!\!\!\!\!\!\!\!\!\!\!\!\!\!\!\!\!\!\geqslant\!\|\bR_{\br_0}\bx_i\!+\!\bt_0\!-\!\by_{j^*}\!\|\!-\!({\gamma_r}_i\!+\!\gamma_t) \label{eq:ineq_uncertaintyradius} \\
&&\!\!\!\!\!\!\!\!\!\!\!\!\!\!\!\!\!\!\geqslant\!\|\bR_{\br_0}\bx_i\!+\!\bt_0\!-\!\by_{j^*_0}\!\|\!-\!({\gamma_r}_i\!+\!\gamma_t) \label{eq:ineq_closestpoint} \\
&&\!\!\!\!\!\!\!\!\!\!\!\!\!\!\!\!\!\!=\!e_i(\bR_{\br_0}, \bt_0)\!-\!({\gamma_r}_i\!+\!\gamma_t),
\end{eqnarray} where (\ref{eq:eq_auxiliary}) trivially involves introducing two auxiliary elements $\bR_{\br_0}\bx$ and $\bt_0$, (\ref{eq:ineq_triangle}) follows from the reverse triangle inequality\footnote{$|x+y| = |x-(-y)| \geqslant |x|-|-y| = |x|-|y|$}, (\ref{eq:ineq_uncertaintyradius}) is based on the uncertainty radii in (\ref{eq:rotuncertainty}) and (\ref{eq:transuncertainty}), and (\ref{eq:ineq_closestpoint}) is from the closest-point definition.  Note that $\by_{j^*}$ is not fixed, but changes dynamically as a function of $(\bR_\br,\bt)$ as defined in (\ref{eq:closestpoint}).

According to the above derivation, the residual error $e_i(\bR_\br, \bt)$ after perturbing a data point $\mathbf{x}_i$ by a 3D rigid motion composed of a rotation $\br\in C_r$ and a translation $\mathbf{t}\!\in\! C_t$ will be at least $e_i(\bR_{\br_0}, \bt_0)\!-\!({\gamma_r}_i\!+\!\gamma_t)$. Given that a closest point distance should be non-negative, a valid lower bound $\underline{e_i}$ for $C_r\times C_t$ is $\max\big(e_i(\bR_{\br_0}, \bt_0)-({\gamma_r}_i\!+\!\gamma_t),0\big)\!\leqslant\! \min_{\forall(\br,\bt)\in(C_r\times C_t)}e_i(\bR_\br,\bt)$.\qed
\vspace{3pt}

\begin{figure}
\begin{center}
\includegraphics[width=0.26\textwidth]{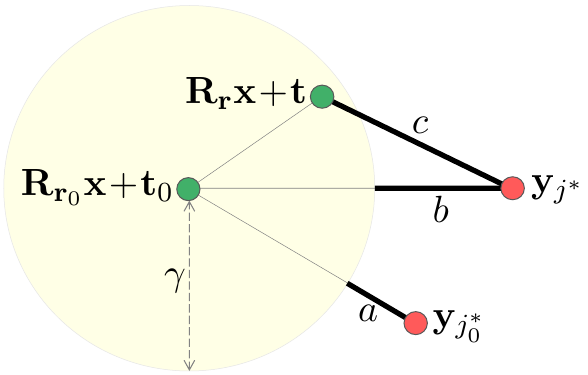}
\vspace{0pt}
\caption{Deriving the lower bound. Any transformed data point $\bR_\br \bx\!+\!\bt$ lies within the uncertainty ball (in yellow) centered at $\bR_{\br_0} \bx\!+\!\bt_0$ with radius $\gamma=\gamma_r+\gamma_t$. Model points $\by_{j^*}$ and $\by_{j^*_0}$ are closest to $\bR_\br \bx\!+\!\bt$ and $\bR_{\br_0} \bx\!+\!\bt_0$ respectively. It is clear that $a\le b \le c$ where $a=\underline{e_i}$ and $c=e_i(\bR_\br,\bt)$. See text for more details.
\label{fig:lowerbound}}
\end{center}
\vspace{0pt}
\end{figure}

The geometric explanation for $\underline{e_i}$ is as follows. Since $\by_{j^*_0}$ is closest to the center $\bR_{\br_0}\bx_i+\bt_0$ of the uncertainty ball with radius ${\gamma={\gamma_r}_i+\gamma_t}$, it is also closest to the surface of the ball and $\underline{e_i}$ is the closest distance between point-set $\mathcal{Y}$ and the ball. Thus, no matter where the transformed data point $\bR_\br \bx_i+\bt$ lies inside the ball, its closest distance to point-set $\mathcal{Y}$ will be no less than $\underline{e_i}$. See Fig.~\ref{fig:lowerbound} for a geometric illustration.

Summing the squared upper and lower bounds of per-point residuals in (\ref{eq:perpointupperbound}) and (\ref{eq:perpointlowerbound}) for all $M$ points, we get the $L_2$-error bounds in the following corollary.

\begin{corollary}\label{cor:l2errorbounds}(Bounds of $L_2$ error) For a 3D motion domain $C_r\times C_t$ centered at $(\br_0,\bt_0)$ with uncertainty radii ${\gamma_r}_i$ and $\gamma_t$, the upper bound $\overline{E}$ and the lower bound $\underline{E}$ of the optimal $L_2$ registration error $E^*$ can be chosen as
\begin{eqnarray}
    &&\!\!\!\!\!\!\!\!\!\!\!\!\!\!\!\!\!\!\!\overline{E} \doteq \sum_{i=1}^M \overline{e_i}^2 = \sum_{i=1}^M e_i(\bR_{\br_0},\!\bt_0)^2, \label{eq:upperbound} \vspace{-5pt}\\
    &&\!\!\!\!\!\!\!\!\!\!\!\!\!\!\!\!\!\!\!\underline{E} \doteq \sum_{i=1}^M \underline{e_i}^2=
    \sum_{i=1}^M \max\big(e_i(\bR_{\br_0},\!\bt_0)\!-\!({\gamma_r}_i\!+\!\gamma_t), 0\big)^2.
    \label{eq:lowerbound}
\end{eqnarray}
\end{corollary}

\section{The Go-ICP Algorithm}\label{sec:algorithm}
Now that the domain parametrization and bounding functions have been specified, we are ready to present the Go-ICP algorithm concretely.

\subsection{Nested BnBs}

Given Corollary~\ref{cor:l2errorbounds}, a direct 6D space BnB (\ie branching each 6D cube into $2^6=64$ sub-cubes and bounding errors for them) seems to be straightforward. However, we find it prohibitively inefficient and memory consuming, due to the huge number of 6D cubes and point-set transformation operations.

Instead, we propose using a nested BnB search structure.  An outer BnB searches the rotation space of $SO(3)$ and solves the bounds and corresponding optimal translations by calling an inner translation BnB. \textcolor[rgb]{0.00,0.00,0.00}{In this way, we only need to maintain two queues with significantly fewer cubes. Moreover, it avoids redundant point-set rotation operations for each rotation region, and takes the advantage that translation operations are computationally much cheaper.}

The bounds for both the BnBs can be readily derived according to Sec.~\ref{sec:errorbound}. In the outer rotation BnB, for a rotation cube $C_r$ the bounds can be chosen as
\begin{eqnarray}
\!\!\!\!&&\overline{E}_r = \min_{\forall \bt \in \mathscr{C}_t}\sum_i e_i(\bR_{\br_0}, \bt)^2,  \label{eq:rotationub}\\
\!\!\!\!&&\underline{E}_r = \min_{\forall \bt \in \mathscr{C}_t}\sum_i \max\big(e_i(\bR_{\br_0}, \bt)-{\gamma_r}_i,0\big)^2, \label{eq:rotationlb}
\end{eqnarray}
where $\mathscr{C}_t$ is the initial translation cube. To solve the lower bound $\underline{E}_r$ in (\ref{eq:rotationlb}) with the inner translation BnB, the bounds for a translation cube $C_t$ can be chosen as
\begin{eqnarray}
\!\!\!\!&&\overline{E}_t =\sum_i \max\big(e_i(\bR_{\br_0}, \bt_0)-{\gamma_r}_i, 0\big)^2, \label{eq:translationub}\\
\!\!\!\!&&\underline{E}_t = \sum_i \max\big(e_i(\bR_{\br_0}, \bt_0)-({\gamma_r}_i+\gamma_t),0\big)^2. \label{eq:translationlb}
\end{eqnarray}
By setting all the rotation uncertainty radii ${\gamma_r}_i$ in (\ref{eq:translationub}) and (\ref{eq:translationlb}) to zero, the translation BnB solves $\overline{E}_r$ in  (\ref{eq:rotationub}).
A detailed description is given in Algorithm~\ref{alg:bnbrt} and Algorithm~\ref{alg:bnbt}.

\vspace{0.06in}
\noindent\textbf{Search Strategy and Stop Criterion.} In both BnBs, we use a best-first search strategy. Specifically, each of the BnBs maintains a priority queue; the priority of a cube is opposite to its lower bound. Once the difference between so-far-the-best error $E^*$ and the lower bound $\underline{E}$ of current cube is less than a threshold $\epsilon$, the BnB stops. Another possible strategy is to set $\epsilon=0$ and terminate the BnBs when the remaining cubes are sufficiently small.

\begin{algorithm}[!tb]
\footnotesize
\caption{Go-ICP -- the Main Algorithm: BnB search for optimal registration in $SE(3)$}\label{alg:bnbrt}
\KwIn{Data and model points; threshold $\epsilon$; initial cubes $\mathscr{C}_r$,\! $\mathscr{C}_t$.
}
\KwOut{
  Globally minimal error $E^*$ and corresponding $\br^*$,\! $\bt^*$.
}
    Put $\mathscr{C}_r$ into priority queue $Q_r$.\\
    Set $E^*=+\infty$.\\

    \Begin
    {
        Read out a cube with lowest lower-bound $\underline{E}_r$ from $Q_r$.\\
        Quit the loop if $E^*\!-\!\underline{E}_r\!<\!\epsilon$. \\
        Divide the cube into 8 sub-cubes.\\
        \ForEach{\textnormal{sub-cube} $C_r$}
        {
            Compute $\overline{E}_r$ for $C_r$ and corresponding optimal $\bt$ by calling Algorithm~\ref{alg:bnbt} with $\br_0$, zero uncertainty radii, and $E^*$. \\
            \If{$\overline{E}_r<E^*$}
            {
                {Run ICP with the initialization $(\br_0, \bt)$.} \label{ln:icprefine1}\\
                Update $E^*$, $\br^*$, and $\bt^*$ with the results of ICP. \label{ln:icprefine2}
            }

            Compute $\underline{E}_r$ for $C_r$ by calling Algorithm~\ref{alg:bnbt} with $\br_0$, $\bgamma_r$ and $E^*$.\\
            \If{$\underline{E}_r\geqslant E^*$}
            {
                Discard $C_r$ and continue the loop;
            }
            Put $C_r$ into $Q_r$.
        }
    }
\end{algorithm}

\begin{algorithm}[!tb]
\footnotesize
\caption{BnB search for optimal translation given rotation}\label{alg:bnbt}
\KwIn{Data and model points; threshold $\epsilon$; initial cube $\mathscr{C}_t$; rotation $\br_0$; rotation uncertainty radii $\bgamma_r$, so-far-the-best error $E^*$.
}
\KwOut{Minimal error $E^*_t$ and corresponding $\bt^*$.}
    Put $\mathscr{C}_t$ into priority queue $Q_t$.\\
    Set $E^*_t=E^*$.\\\label{ln:initalerror}
    \Begin
    {
        Read out a cube with lowest lower-bound $\underline{E}_t$ from $Q_t$.\\
        Quit the loop if $E^*_t\!-\!\underline{E}_t\!<\!\epsilon$. \\
        Divide the cube into 8 sub-cubes.\\
        \ForEach{\textnormal{sub-cube} $C_t$}
        {
            Compute $\overline{E}_t$ for $C_t$ by (\ref{eq:translationub}) with $\br_0$,$\bt_0$ and $\bgamma_r$.\\
            \If{$\overline{E_t}<E^*_t$}
            {
                Update $E^*_t=\overline{E}_t$, $\bt^*=\bt_0$.\\
            }
            Compute $\underline{E}_t$ for $C_t$ by (\ref{eq:translationlb}) with $\br_0$,$\bt_0$,$\bgamma_r$ and $\gamma_t$.\\
            \If{$\underline{E}_t\geqslant E^*_t$}
            {
                Discard $C_t$ and continue the loop.\\
            }
            Put $C_t$ into $Q_t$.
        }
    }
\end{algorithm}

\subsection{Integration with the ICP Algorithm}\label{sec:integration}
Lines~\ref{ln:icprefine1}--\ref{ln:icprefine2} of Algorithm~\ref{alg:bnbrt} show that whenever the outer BnB finds a cube $C_r$ that has an upper bound lower than the current best function value, it will call conventional ICP, initialized with the center rotation of $C_r$ and the corresponding best translation.

\begin{figure}
\begin{center}
\includegraphics[width=0.475\textwidth]{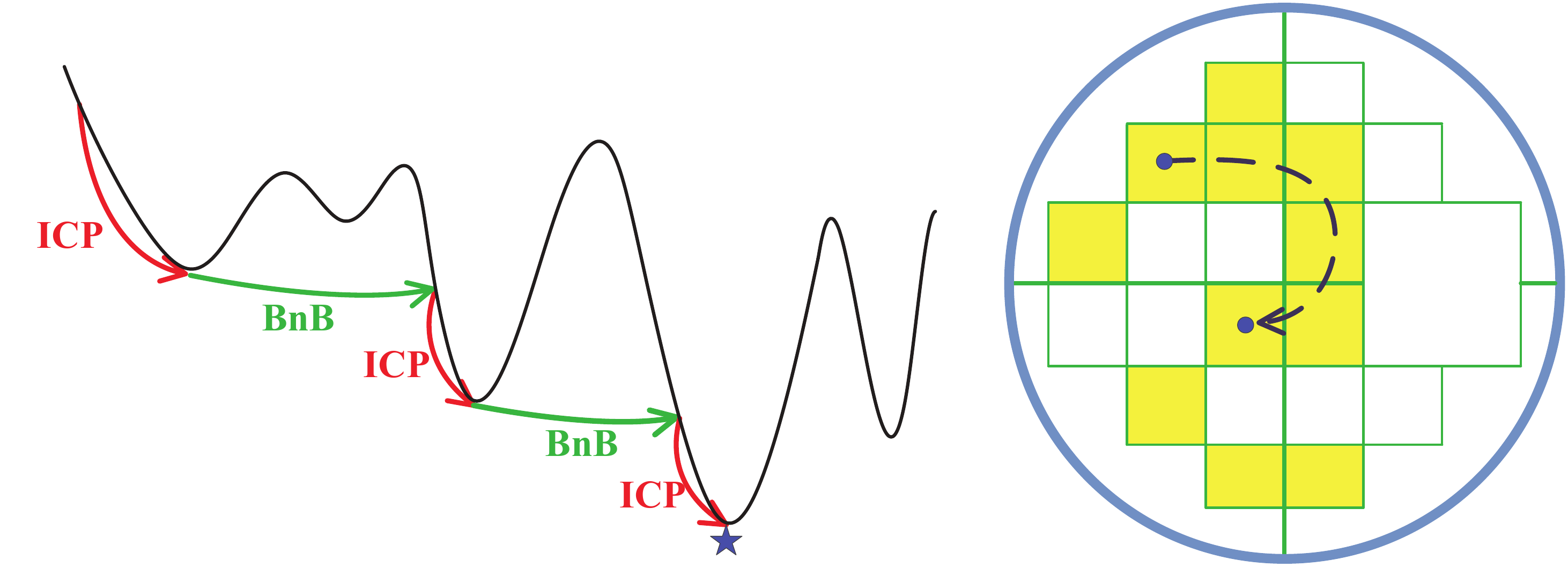}
\caption{Collaboration of BnB and ICP. \textbf{Left}: BnB and ICP collaboratively update the upper bounds during the search process.  \textbf{Right}: with the guidance of BnB, ICP only explores un-discarded, promising cubes with small lower bounds marked up by BnB. \label{fig:ICP_BnB_coop}}
\end{center}
\vspace{-5pt}
\end{figure}

Figure~\ref{fig:ICP_BnB_coop} illustrates the collaborative relationship between ICP and BnB. Under the guidance of global BnB, ICP converges into local minima one by one, with each local minimum having lower error than the previous one, and ultimately reaches the global minimum.
Since ICP monotonically decreases the current-best error $E^*$ (cf. \cite{besl1992method}), the search path of the local ICP is confined to un-discarded, promising sub-cubes with small lower bounds, as illustrated in Fig.~\ref{fig:ICP_BnB_coop}.

In this way, the global BnB search and the local ICP search are intimately integrated in the proposed method. The former helps the latter jump out of local minima and guides the latter's next search; the latter accelerates the former's convergence by refining the upper bound, hence improving the efficiency.

\subsection{Outlier Handling with Trimming}\label{sec:outlier}
\begin{figure*}[!t]
\begin{center}
\includegraphics[width=1\textwidth]{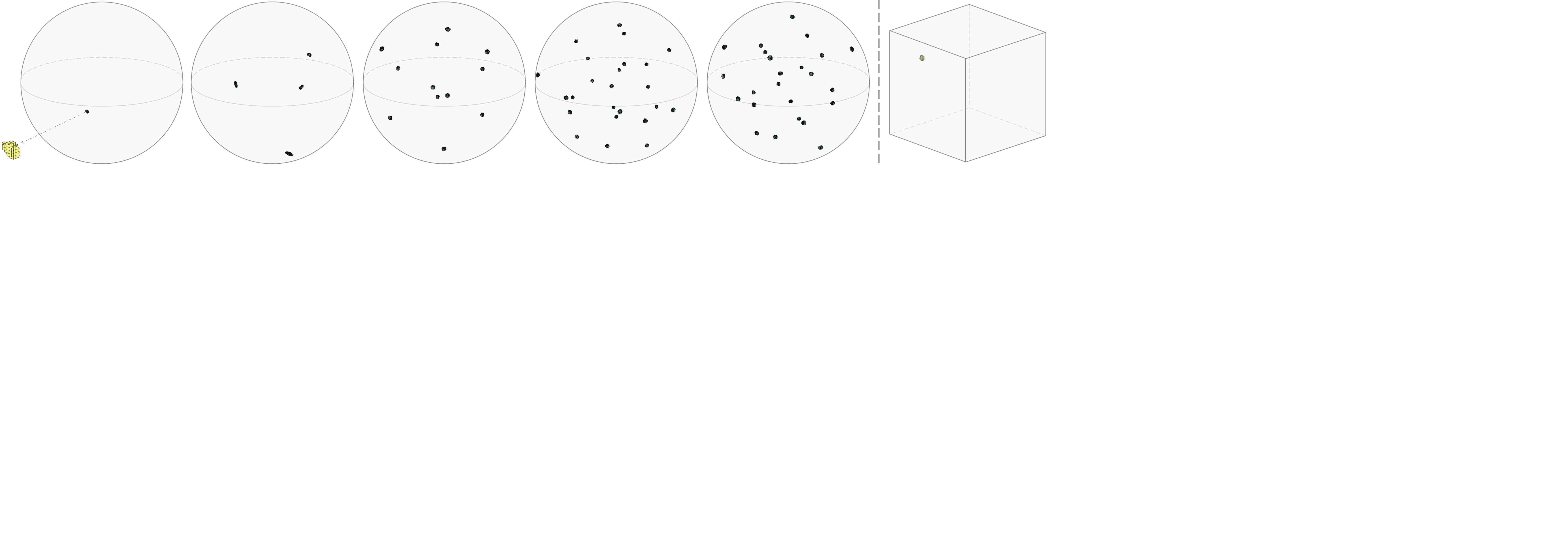}
\caption{Remaining cubes of the BnBs. The first five figures show the remaining cubes in the rotation $\pi$-ball of the rotation BnBs, for an irregular tetrahedron, a cuboid with three different side-lengths, a regular tetrahedron, a regular cube, and a regular octahedron respectively. The last figure shows a typical example of remaining cubes of a translation BnB, for the irregular tetrahedron. (Best viewed when zoomed in)
\label{fig:rotationcubes}}
\end{center}
\end{figure*}

In statistics, trimming is a strategy to obtain a more robust statistic by excluding some of the extreme values. It is used in Trimmed ICP~\cite{chetverikov2005robust} for robust point-set registration. Specifically, in each iteration, only a subset $\mathcal{S}$ of the data points that have the smallest closest distances are used for motion computation. Therefore, the registration error will be
\begin{equation}\label{eq:trimmedregistrationerror}
    E^{Tr}=\sum_{i\in\mathcal{S}}e_i(\bR,\bt)^2.
\end{equation}
To robustify our method with trimming, it is necessary to derive new upper and lower bounds of (\ref{eq:trimmedregistrationerror}). We have the following result.
\begin{corollary}
    (Bounds of the trimmed $L_2$ error) The upper bound $\overline{E^{Tr}}$ and lower bound $\textcolor[rgb]{0.00,0.00,0.00}{\underline{E^{Tr}}}$ of the registration error with trimming for the domain $C_r\times C_t$  can be chosen as
    \begin{eqnarray}
        &&\overline{E^{Tr}}\doteq\sum_{i\in\mathcal{P}}\overline{e_i}^2, \label{eq:trimmedperpointupperbound}\\
        &&\underline{E^{Tr}}\doteq\sum_{i\in \mathcal{Q}}\underline{e_i}^2.
        \label{eq:trimmedperpointlowerbound}
    \end{eqnarray}
    where $\overline{e_i}$, $\underline{e_i}$ are bounds of the per-point residuals defined in (\ref{eq:perpointupperbound}), (\ref{eq:perpointlowerbound}) respectively, and $\mathcal{P}$, $\mathcal{Q}$ are the trimmed point-sets having smallest values of $\overline{e_i}$, $\underline{e_i}$ respectively, with ${|\mathcal{P}|=|\mathcal{Q}|=|\mathcal{S}|=K}$.
\end{corollary}
\noindent\emph{Proof:}
The upper bound in (\ref{eq:trimmedperpointupperbound}) is chosen trivially.
To see the validity of the lower bound in (\ref{eq:trimmedperpointlowerbound}), observe that $\forall (\br,\bt)\in C_r\times C_t$,
\begin{equation}
\underline{E^{Tr}}=\sum_{i\in \mathcal{Q}}\underline{e_i}^2 \leq \sum_{i\in \mathcal{S}}\underline{e_i}^2 \leq \sum_{i\in \mathcal{S}}e_i(\bR_\br,\bt)^2 = E^{Tr}\!.\!
\end{equation}\qed

Based on this corollary, the corresponding bounds in the nested BnB can be readily derived. As proved in \cite{chetverikov2005robust}, iterations of Trimmed ICP decrease the registration error monotonically to a local minimum. Thus it can be directly integrated into the BnB procedure.

\vspace{0.06in}
\noindent\textbf{Fast Trimming.}
A straightforward yet inefficient way to do trimming is to sort the residuals outright and use the $K$ smallest ones. In this paper, we employ \textcolor[rgb]{0.00,0.00,0.00}{the Introspective Selection algorithm}~\cite{musser1997introspective} which has $O(N)$ performance in both the worst case and average case.

\vspace{0.06in}
\noindent\textbf{Other Robust Extensions.}
In the same spirit as trimming, other ICP variants such as \cite{champleboux1992accurate,masuda1994robust} can be handled. The method can also be adapted to LM-ICP~\cite{fitzgibbon2003robust}, where the new lower-bound is simply a robust kernelized version of the current one. It may also be extended to ICP variants with $L_p$-norms~\cite{bouaziz2013sparse}, such as the robustness-promoting $L_1$-norm.

\section{Experiments}\label{sec:experiment}
We implemented the method\footnote{Source code and demo can be found on the author's webpage.} in C++ and tested it on a standard PC with an Intel i7 3.4GHz CPU. In the experiments reported below, the point-sets were pre-normalized such that all the points were within the domain of $[-1,1]^3$. Although the goal was to minimize the $L_2$ error in (\ref{eq:registrationerror}), the root-mean-square (RMS) error is reported for better comprehension.

\vspace{0.06in}
\noindent\textbf{Closest-point distance computation.} To speed up the closest distance computation, a kd-tree data structure can be used. We also provide an alternative solution that is used more often in the experiments -- a 3D Euclidean Distance Transform (DT)~\cite{fitzgibbon2003robust} used to compute closest distances for fast bound evaluation\footnote{Local ICP is called infrequently so we simply use a kd-tree for it. The refined upper-bounds from the found ICP solutions are evaluated via the DT for consistency.
}.
A DT approximates the closest-point distances in the real-valued space by distances of uniform grids, and pre-computes them for constant-time retrieval (details about our DT implementation can be found in the supplementary material). Despite the DT can introduce approximation errors thus the convergence gap may not be exactly $\epsilon$, in the following experiments our method works very well with a $300\!\times\!300\!\times\!300$ DT for optimal registration. Naturally, higher resolutions can be used when necessary.

\begin{figure}[!t]
\begin{center}
\includegraphics[width=0.425\textwidth]{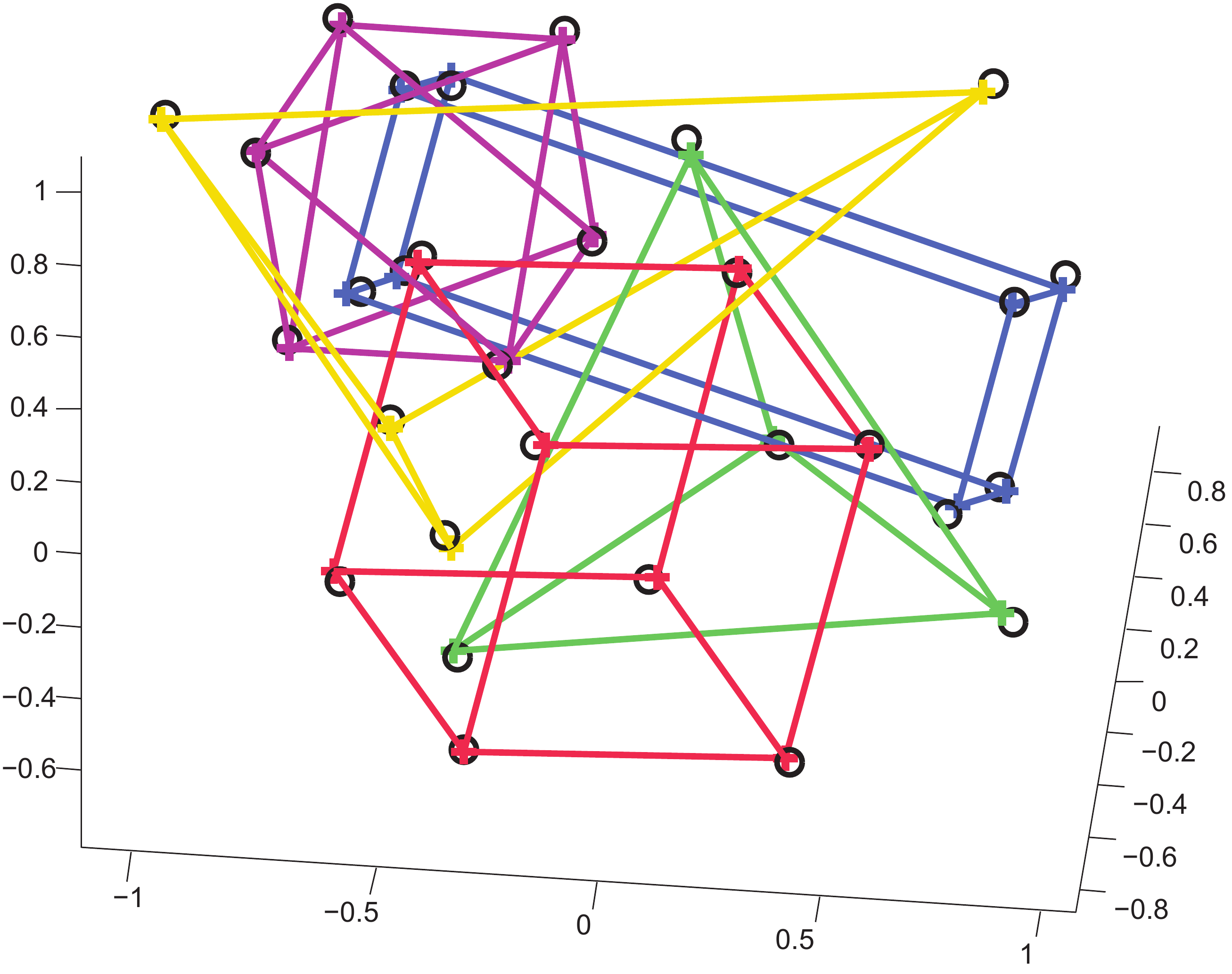}
\caption{A clustered scene (black circles) and the registration results of Go-ICP for the five shapes.
\label{fig:syntheticscene}}
\end{center}
\vspace{-5pt}
\end{figure}

\begin{figure}[!t]
\begin{center}
\includegraphics[width=0.49\textwidth]{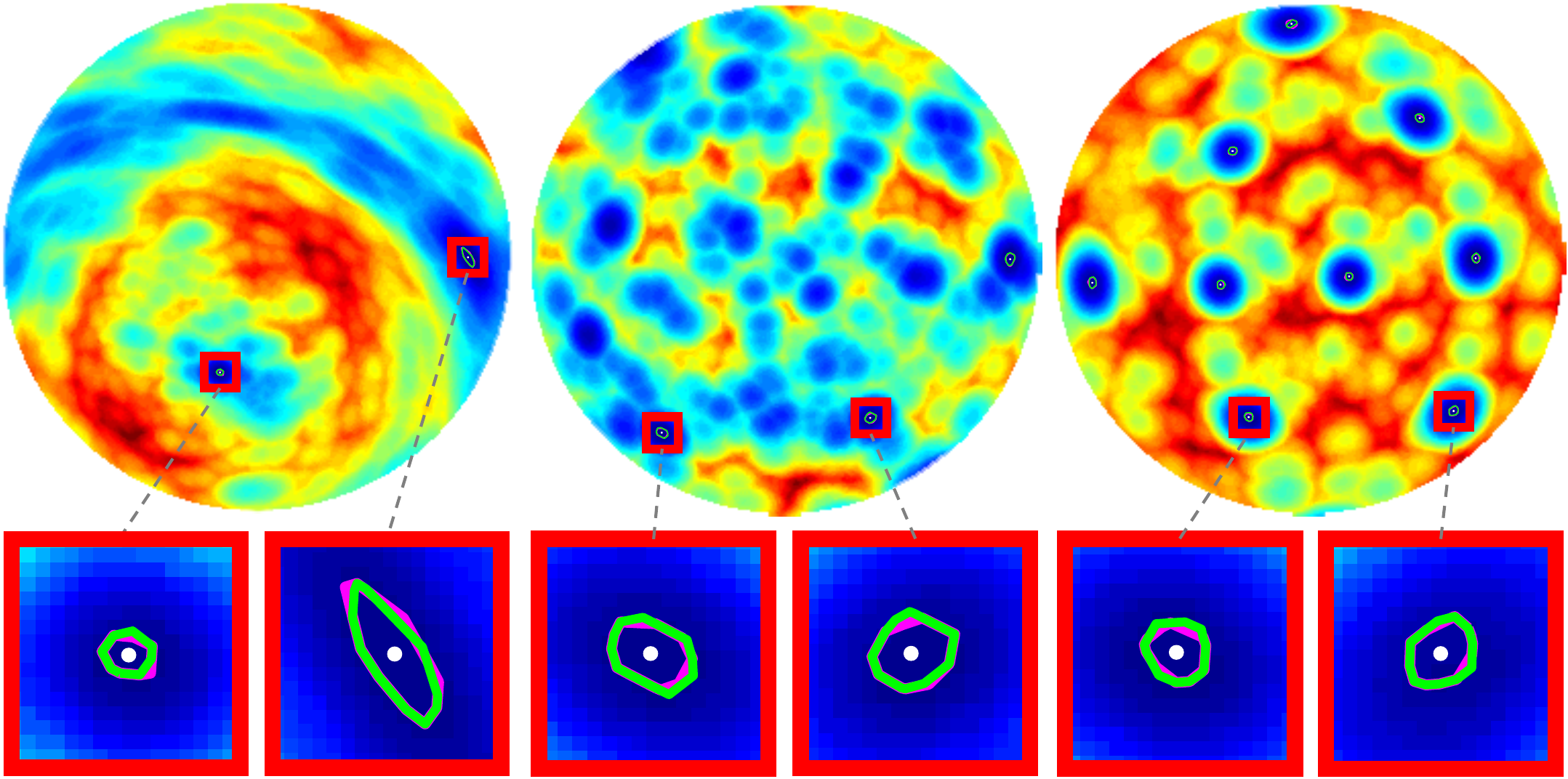}
\caption{Remaining rotation domains of the outer rotation BnB on 2D slices of the $\pi$-ball, for the synthetic points. Results using the DT and the kd-tree are within magenta and green polygons, respectively. The white dots denote optimal rotations. From left to right: a cuboid, a regular tetrahedron and a regular cube. The colors on the slices indicate registration errors evaluated via inner translation BnB: red for high error and blue for low error. (Best viewed when zoomed in)
\label{fig:syntheticrotationslice}}
\end{center}
\vspace{-5pt}
\end{figure}

\begin{figure*}[!t]
\begin{center}
\includegraphics[width=0.995\textwidth]{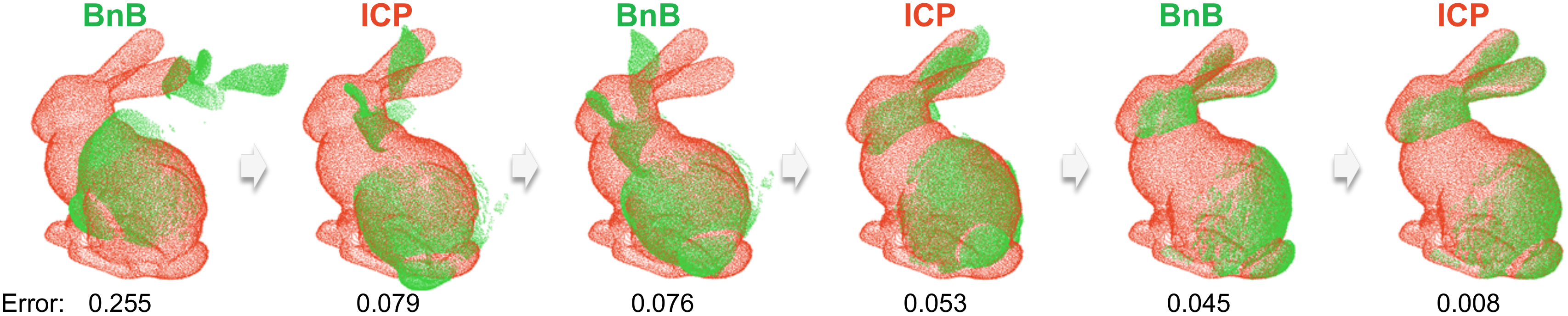}
\caption{Evolution of Go-ICP registration for the bunny dataset. The model point-set and data point-set are shown in red and green respectively. BnB and ICP collaboratively update the registration: ICP refines the solution found by BnB and BnB guides ICP into the convergence basins of multiple local minima with increasingly lower registration errors.
\label{fig:regevol}}
\end{center}
\vspace{-5pt}
\end{figure*}

\subsection{Optimality}\label{sec:optimality}
To verify the correctness of the derived bounds and the optimality of Go-ICP, we first use a convergence condition similar to \cite{hartley2009global} for the BnBs. Specifically, we set the the threshold of a BnB to be 0, and specify a smallest cube size at which the BnB stops dividing a cube. In this way, we can examine the uncertainty in the parameter space after the BnB stops. Both the DT and kd-tree are tested in these experiments.

\vspace{0.06in}
\noindent\textbf{Synthetic Points.}
We first tested the method on a synthetically generated scene with simple objects. Specifically, five 3D shapes were created: an irregular tetrahedron, a cuboid with three different side-lengths, a regular tetrahedron, a regular cube, and a regular octahedron. Note that the latter 4 shapes have self-symmetries. All the shapes were then placed together, each with a random transformation, to generate clustered scenes. Zero-mean Gaussian noise with standard deviation $\sigma\!=\!0.01$ was added to the scene points. We created such a scene as shown in Figure~\ref{fig:syntheticscene}, and applied Go-ICP to register the vertices of each shape to the scene points.

To test the rotation BnB, we set the parameter domain to be $[-\pi,\pi]^3\times[-1,1]^3$ and the minimal volume of a rotation cube to $1.5\mathrm{E}{-5}$ ($\sim\!\!1$ degree uncertainty). The lower bound of a rotation cube was set to be the global lower bound of the invoked translation BnB. Thus the threshold of translation BnB is not very important and we set it to a small value ($0.0001\!\times\!N$ where $N$ is the data point number). The initial errors $E^*_t$ of translation BnBs were set to infinity.

In all tests, Go-ICP produced correct results with both the DT and kd-tree. The remaining rotation cubes using the DT and kd-tree respectively are almost visually indistinguishable, and Figure~\ref{fig:rotationcubes} shows the results using the DT. It is interesting to see that the remaining cubes formed 1 cluster for the irregular tetrahedron, 4 clusters for the cuboid, 12 clusters for the regular tetrahedron, and 24 clusters for the regular cube and octahedron. These results conform to the geometric properties of these shapes, and validated the derived bounds. Investigating shape self-similarity would be a practical application of the algorithm.

Moreover, Figure~\ref{fig:syntheticrotationslice} shows some typical remaining rotation domains on 2D slices of the rotation $\pi$-ball\footnote{We chose the slices passing two randomly-selected optimal rotations plus the origin. Due to shape symmetry there may exist more than two optimal rotations on one slice.}. The non-convexity of the problem can be clearly seen from the presence of many local minima. It can also been seen that the remaining rotation domains using a DT and kd-tree are highly consistent, and the optima are well contained by them.

The translation BnB can be easily verified by running it with rotations picked from the remaining rotation cubes. The threshold was set to be 0, and the minimal side-length of a translation cube was set to be $0.01$. The last figure of Fig.~\ref{fig:rotationcubes} shows a typical result.

More results of remaining rotation and translation cubes can be found in the supplementary material.

\begin{figure}[!t]
\begin{center}
\subfigure{
\includegraphics[width=0.2185\textwidth]{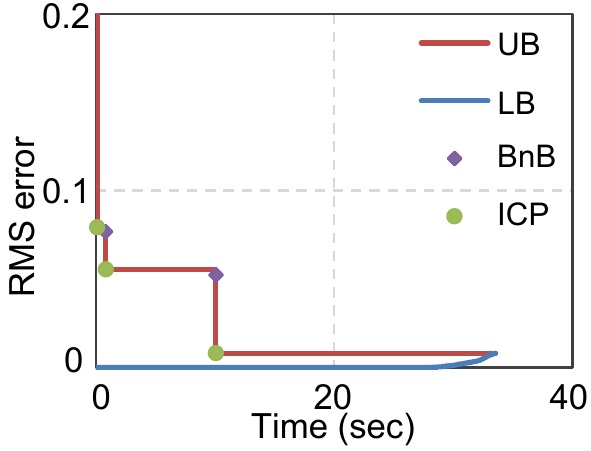}}
\subfigure{
\includegraphics[width=0.244\textwidth]{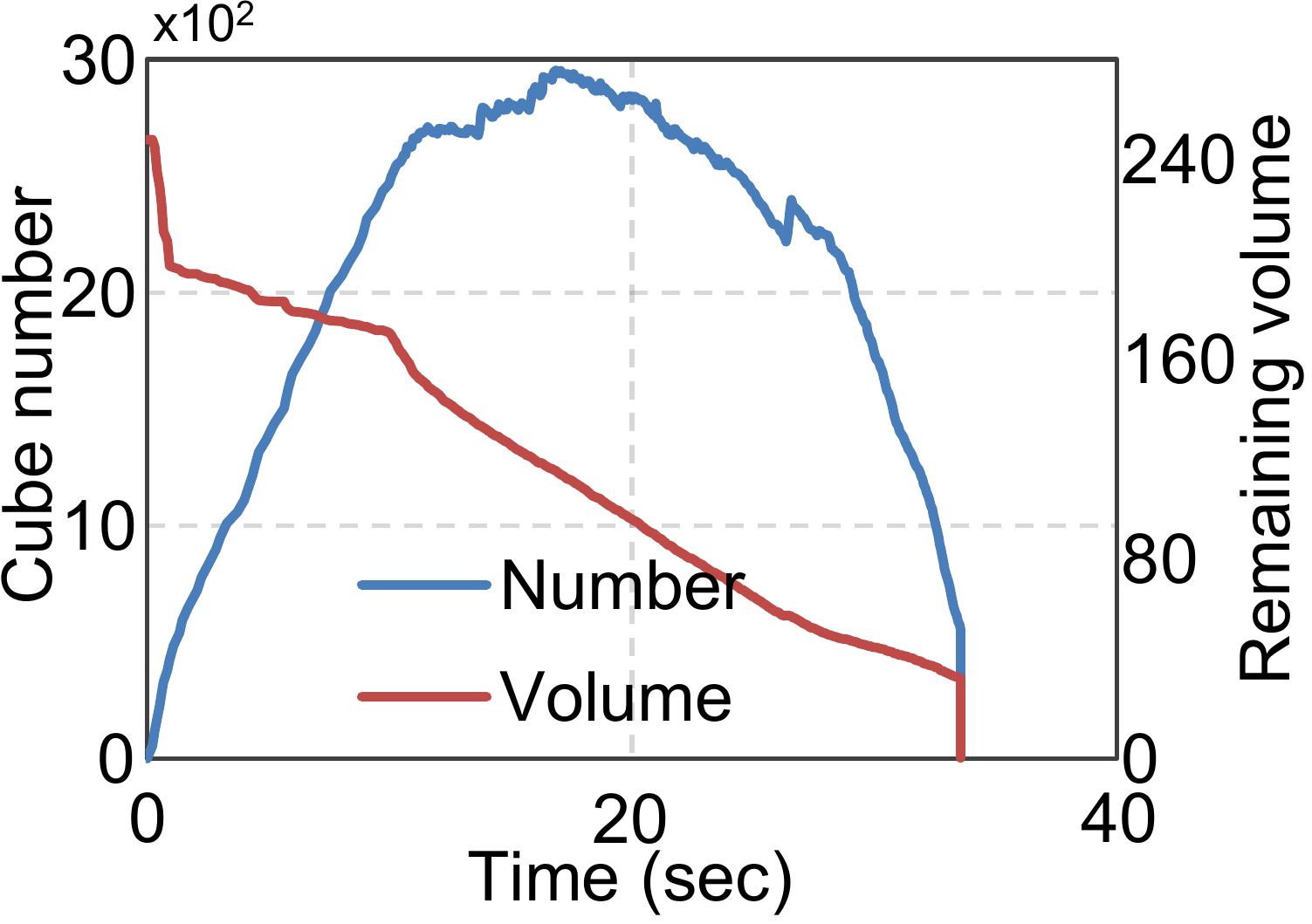}}
\vspace{-13pt}
\caption{Evolution of the bounds (left) and cubes (right) in the rotation BnB with a DT on the bunny point-sets. See text for details.
\label{fig:cubeevol}}
\end{center}
\vspace{-7pt}
\end{figure}

\begin{figure}[!t]
\begin{center}
\includegraphics[width=0.49\textwidth]{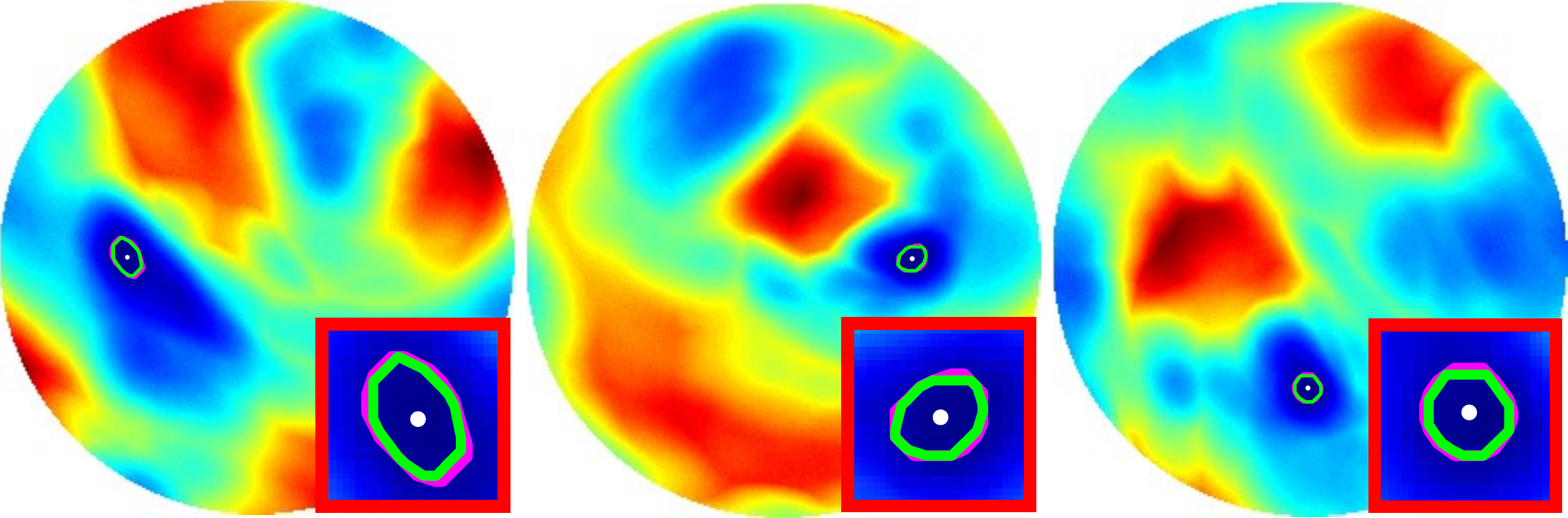}
\caption{Remaining rotation domains of the outer rotation BnB on 2D slices of the $\pi$-ball, for the bunny point-sets. The three slices pass through the optimal rotation and the X-, Y-, Z-axes respectively. See also the caption of Fig.~\ref{fig:syntheticrotationslice}. (Best viewed when zoomed in)
\label{fig:bunnyrotationslice}}
\end{center}
\vspace{-7pt}
\end{figure}

\begin{figure*}[!t]
\begin{center}
\includegraphics[width=1\textwidth]{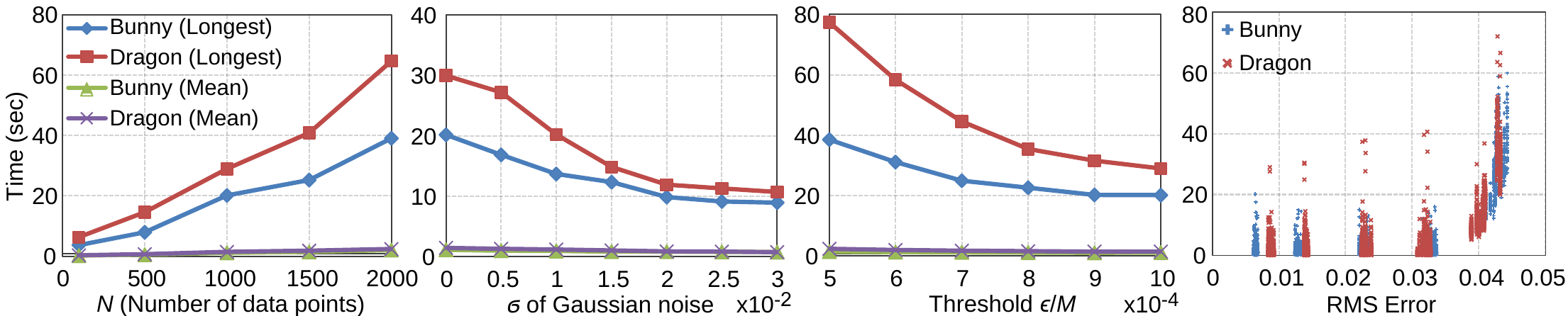}
\caption{Running time of the Go-ICP method with DTs on the bunny and dragon point-sets with respect to different factors. The evaluation was conducted on 10 data point-sets with 100 random poses (\ie, 1\,000 pairwise registrations).
\label{fig:time}}
\end{center}
\vspace{-8pt}
\end{figure*}

\vspace{0.06in}
\noindent\textbf{Real Data.} Similar experiments were conducted on real data. We applied our method to register a bunny scan bun090 from the Stanford 3D dataset\footnote{\url{http://graphics.stanford.edu/data/3Dscanrep/}} to the reconstructed model. Since model and data point-sets are of similar spatial extents, we set the parameter domain to be $[-\pi,\pi]^3\!\times\![-0.5,0.5]^3$ which is large enough to contain the optimal solution. We randomly sampled 500 data points, and did similar tests to those on the synthetic points. The translation BnB threshold was set to $0.001\!\times\!N$, and the remaining rotation cubes from the outer rotation BnB were similar to the first figure in Fig.~\ref{fig:rotationcubes} (\ie, one cube cluster). Figure~\ref{fig:bunnyrotationslice} shows the results on three slices of the rotation $\pi$-ball.

Additionally, we recorded the bound and cube evolutions in the rotation BnB which are presented in Fig.~\ref{fig:cubeevol}. It can be seen that BnB and ICP collaboratively update the global upper bound. Corresponding transformations for each global upper bound found by BnB and ICP are shown in Fig.~\ref{fig:regevol}. Note that in the fourth image the pose has been very close to the optimal one, which indicates that ICP may fail even if reasonably good initialization is given.

Although the convergence condition used in this section worked successfully, we found that using a small threshold $\epsilon$ of the bounds to terminate a BnB also works well in practice. It is more efficient and produces satisfactory results. In the following experiments, we used this strategy for the BnBs.

\subsection{``Partial" to ``Full" Registration}\label{sec:p2f}
\begin{figure}[!t]
\begin{center}
\subfigure{
\includegraphics[width=0.232\textwidth]{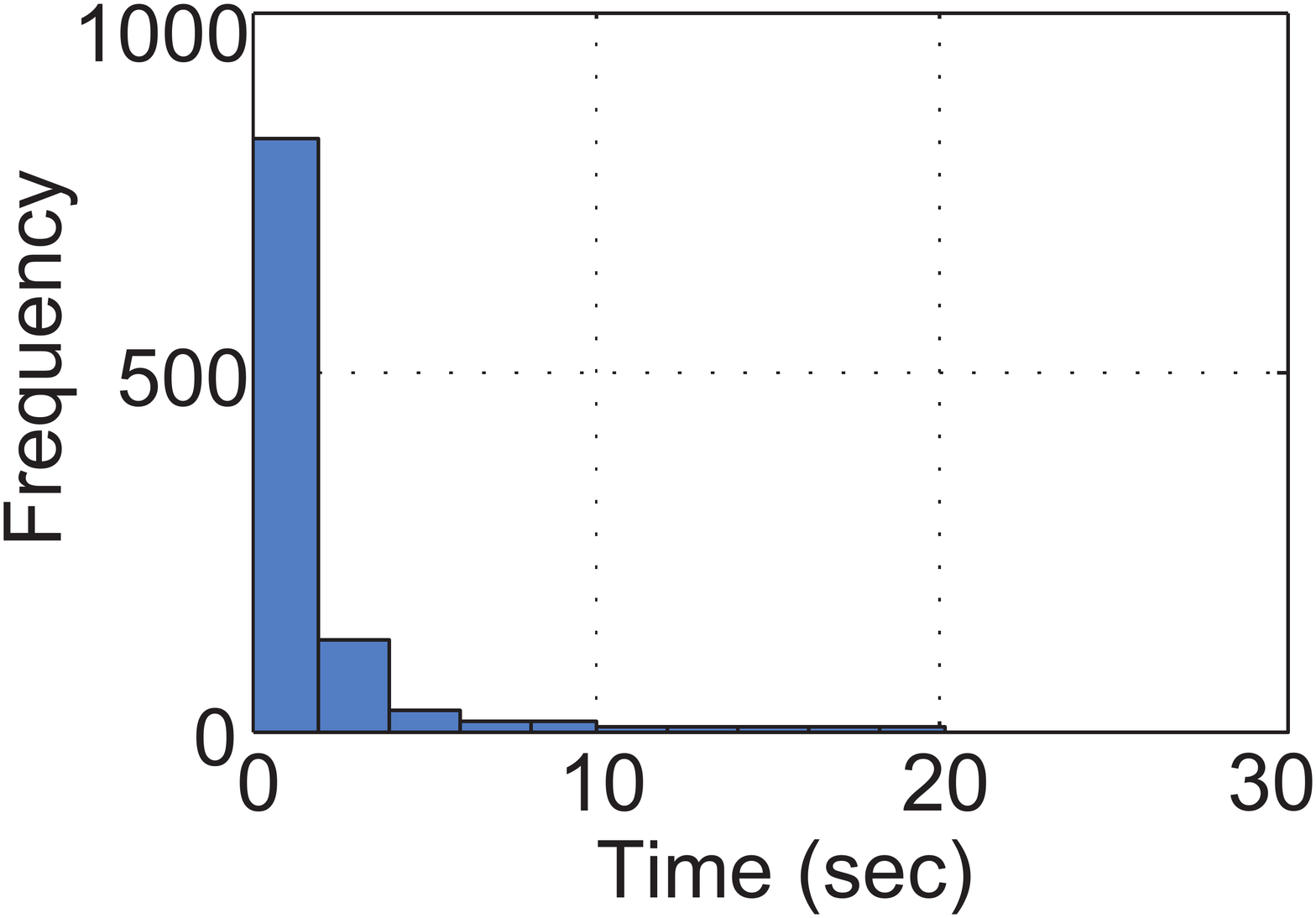}}
\subfigure{
\includegraphics[width=0.232\textwidth]{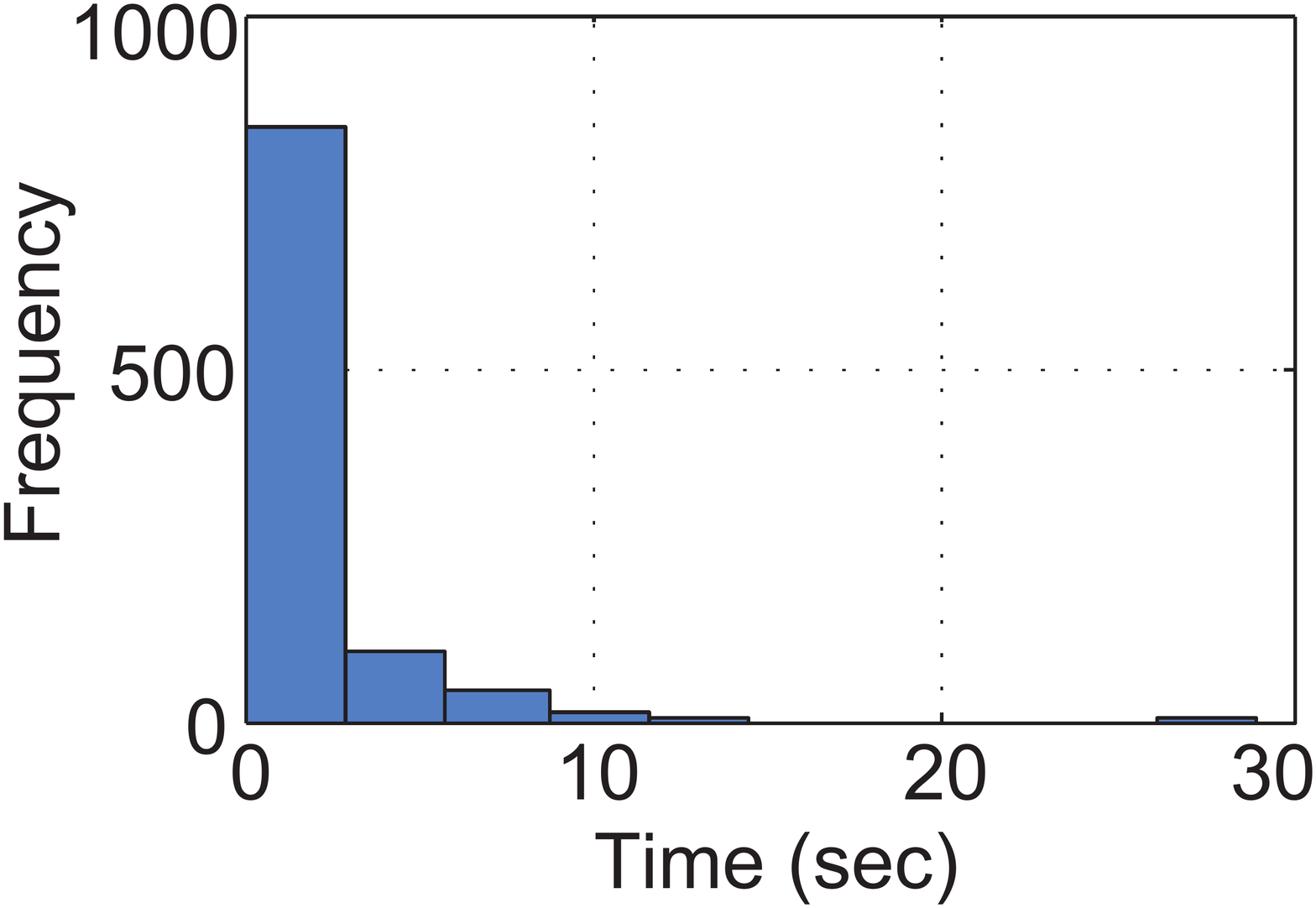}}
\vspace{-10pt}
\caption{Running time histograms of Go-ICP with DTs for the bunny (left) and dragon (right) point-sets.
\label{fig:time_pose}}
\end{center}
\vspace{-9pt}
\end{figure}

In this section, we test the performance of Go-ICP by registering partially scanned point clouds to full 3D model point clouds. The bunny and dragon models from the Stanford 3D dataset were used for evaluation. All 10 partial scans of the bunny dataset were used as data point-sets. For the dragon model, we selected 10 scans generated from different view points as data point-sets. The reconstructed bunny and dragon models were used as model point-sets.

For each of these 20 scans, we first performed 100 tests with random initial rotations and translations. The transformation domain to explore for Go-ICP was set to be $[-\pi,\pi]^3\!\times\![-0.5,0.5]^3$. We sampled $N=1000$ data points from each scan, and set the convergence threshold $\epsilon$ to be $0.001\!\times\!N$.

As expected, \emph{Go-ICP achieved 100\% correct registration on all the 2\,000 registration tasks on the bunny and dragon models}, with both the DT and kd-tree. All rotation errors were less than 2 degrees and all translation errors were less than 0.01. With a DT, the mean/longest running times of Go-ICP, in the 1\,000 tests on 1\,000 data points and 20\,000--40\,000 model points, were 1.6s/22.3s for bunny and 1.5s/28.9s for dragon. Figure~\ref{fig:time_pose} shows the running time histograms The running times with a kd-tree were typically 40--50 times longer than that with the DT. The solutions from using the DT and the kd-tree respectively were highly consistent (the largest rotation difference was below 1 degree). See the supplementary material for detailed result and running time comparisons for the DT and the kd-tree.

We then analyzed the running time of the proposed method under various settings using the DT. We analyzed the influence of each factor by varying it while keeping others fixed. Default factor settings: number of data points $N\!=\!1000$, no added Gaussian noise (\ie standard deviation $\sigma\!=\!0$) and convergence threshold $\epsilon\!=\!0.001\!\times\!N$.

\vspace{0.00in}
\noindent\textbf{Effect of Number of Points.} In this experiment, the running time was tested for different numbers of points. Since the DT was used for closest-point distance retrieval, the number of model points does not significantly affect the speed of our method. To test the running time with respect to different numbers of data points, we randomly sampled the data point-set. As presented in Fig.~\ref{fig:time}, the running time manifested a linear trend since closest-point distance retrieval was $O(1)$ and the convergence threshold varied linearly with the number of data points.

\begin{figure}[!t]
\begin{center}
\captionsetup[subfigure]{labelformat=empty}
\subfigure[$\sigma=0.01$]{
\includegraphics[width=0.16\textwidth]{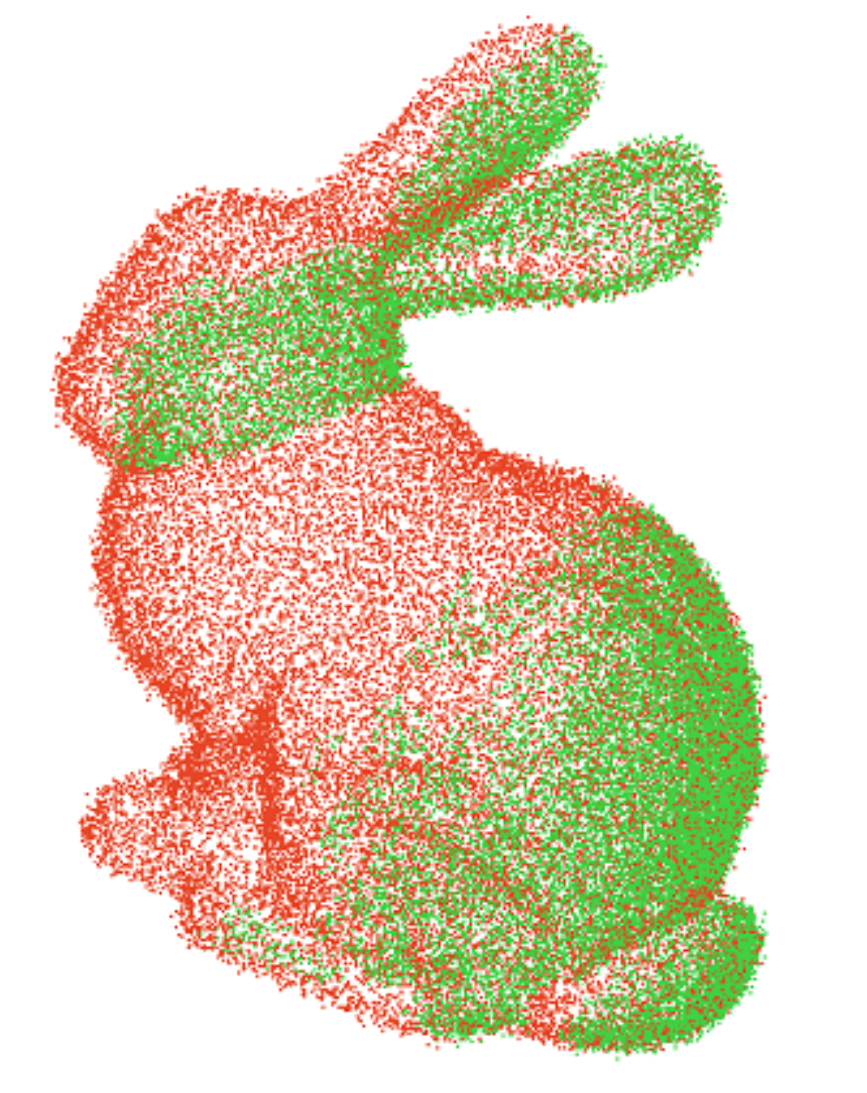}}\!\!\!
\subfigure[$\sigma=0.02$]{
\includegraphics[width=0.16\textwidth]{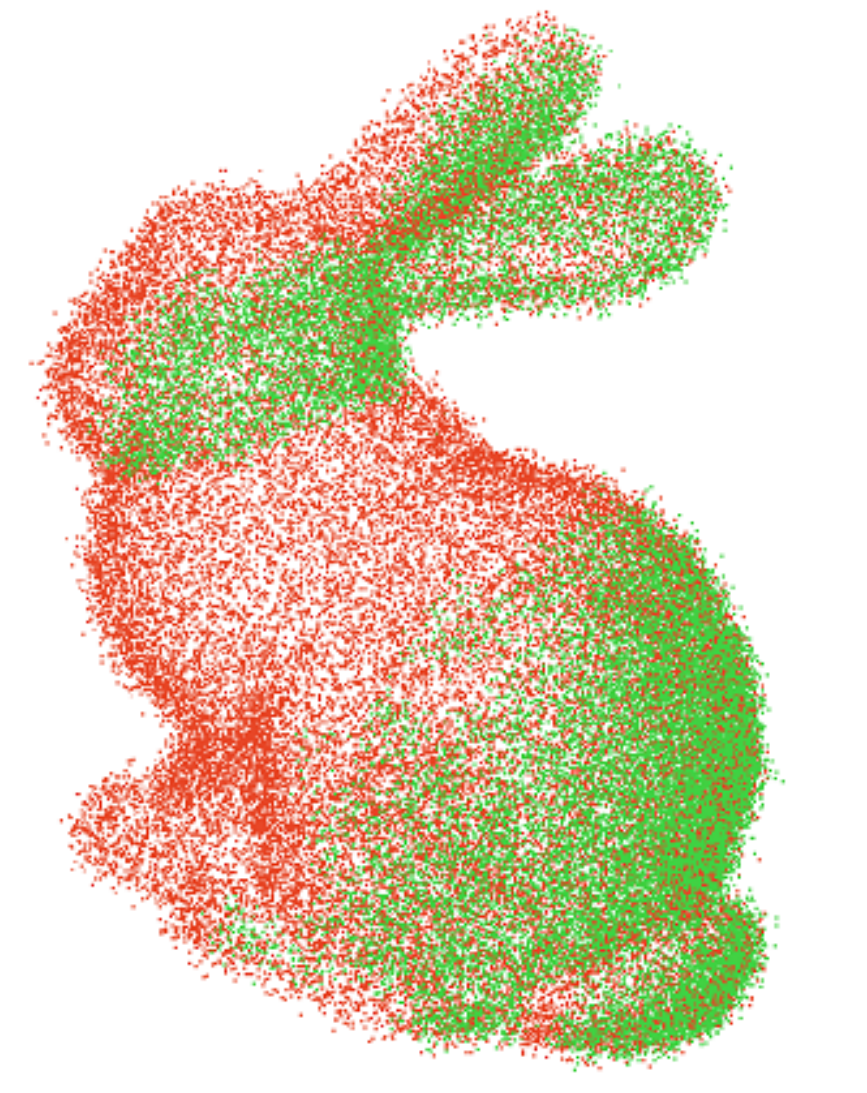}}\!\!\!
\subfigure[$\sigma=0.03$]{
\includegraphics[width=0.16\textwidth]{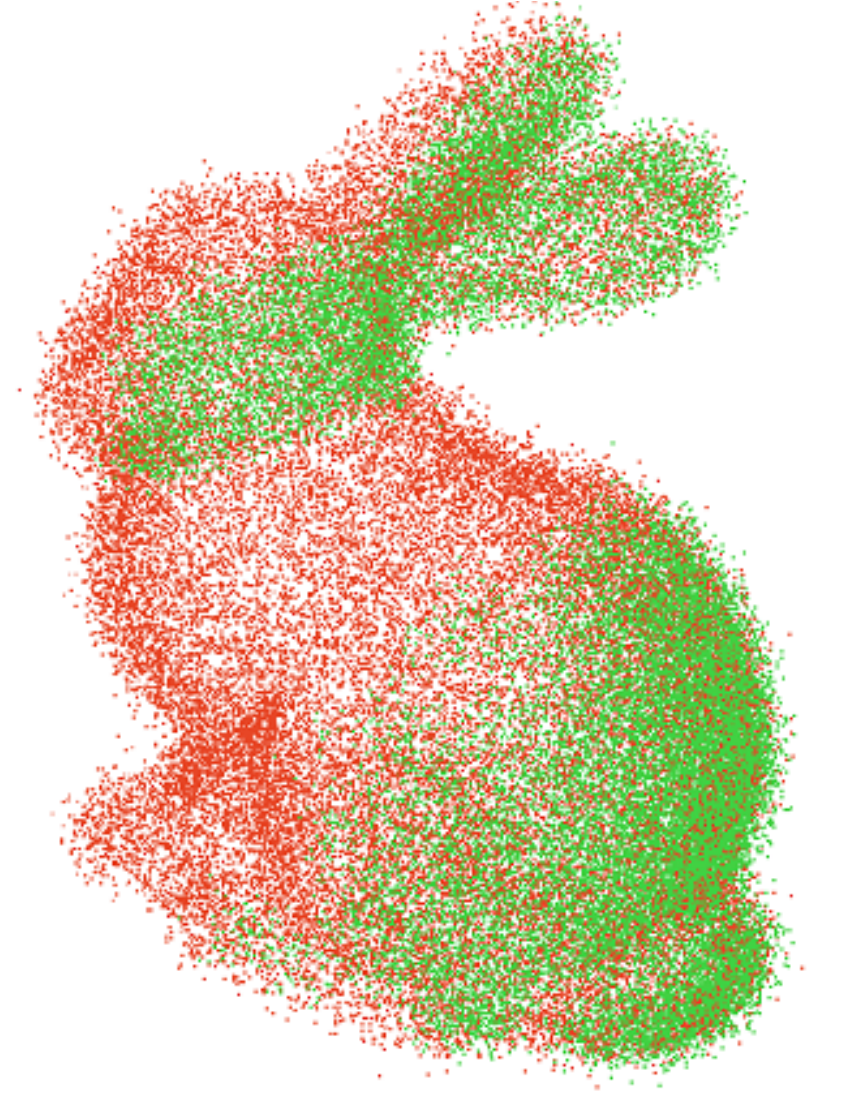}}\!\!\!
\vspace{-4pt}
\caption{Registration with different levels of Gaussian noise.
\label{fig:reg_noise}}
\end{center}
\vspace{-9pt}
\end{figure}

\vspace{0.05in}
\noindent\textbf{Effect of Noise.} We examined how the noise level impacted the running time by adding Gaussian noise to both the data and model point-sets. The registration results on the corrupted bunny point-sets are shown in Fig.~\ref{fig:reg_noise}. We found that, as shown in Fig.~\ref{fig:time},  the running time decreased as the noise level increased (until $\sigma\!=\!0.02$). This is because the Gaussian noise (especially that added to the model points) smoothed out the function landscape and widened the convergence basin of the global minimum, which made it easier for Go-ICP to find a good solution.

\renewcommand*{\thesubfigure}{}
\renewcommand*{\thesubfigure}{(\alph{subfigure})}

\begin{figure*}[!t]
\begin{center}
\includegraphics[width=0.98\textwidth]{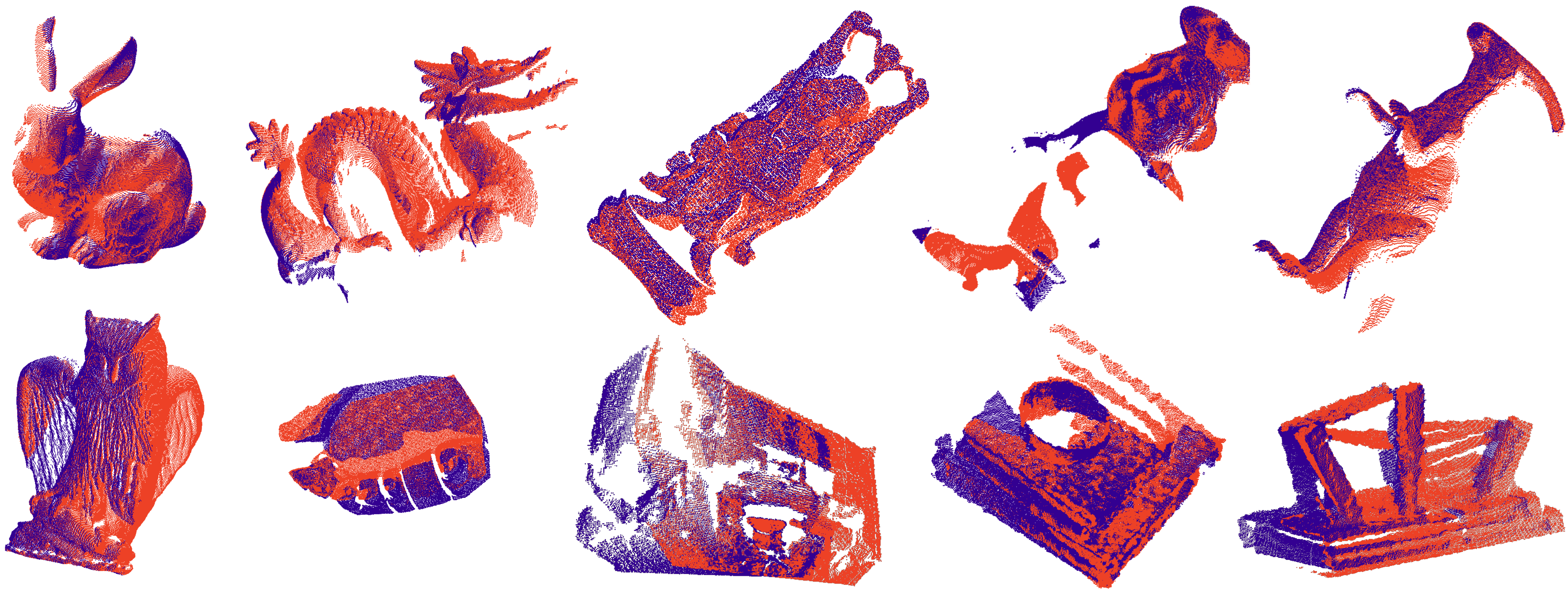}
\caption{Registration with partial overlap. Go-ICP with the trimming strategy successfully registered the 10 point-set pairs with 100 random relative poses for each of them. The point-sets in red and blue are denoted as point-set $A$ and point-set $B$, respectively. The trimming settings and running times are presented in Table~\ref{tab:time_partial}.
\label{fig:partialoverlap}}
\end{center}
\vspace{-2pt}
\end{figure*}

\vspace{0.05in}
\noindent\textbf{Effect of Convergence Threshold.} We further investigated the running time with respect to the convergence threshold of the BnB loops. We set the threshold $\epsilon$ to depend linearly on $N$, since the registration error is a sum over the $N$ data points. Figure~\ref{fig:time} shows that the smaller the threshold is, the slower our method performs. In our experiments, $\epsilon\!=\!0.001\!\times\!N$ was adequate to get a 100\% success rate for the bunny and dragon point-sets. For cases when the local minima are small or close to the global minimum, the threshold can be set smaller.

\begin{figure}[!t]
\begin{center}
\subfigure{
\includegraphics[width=0.136\textwidth]{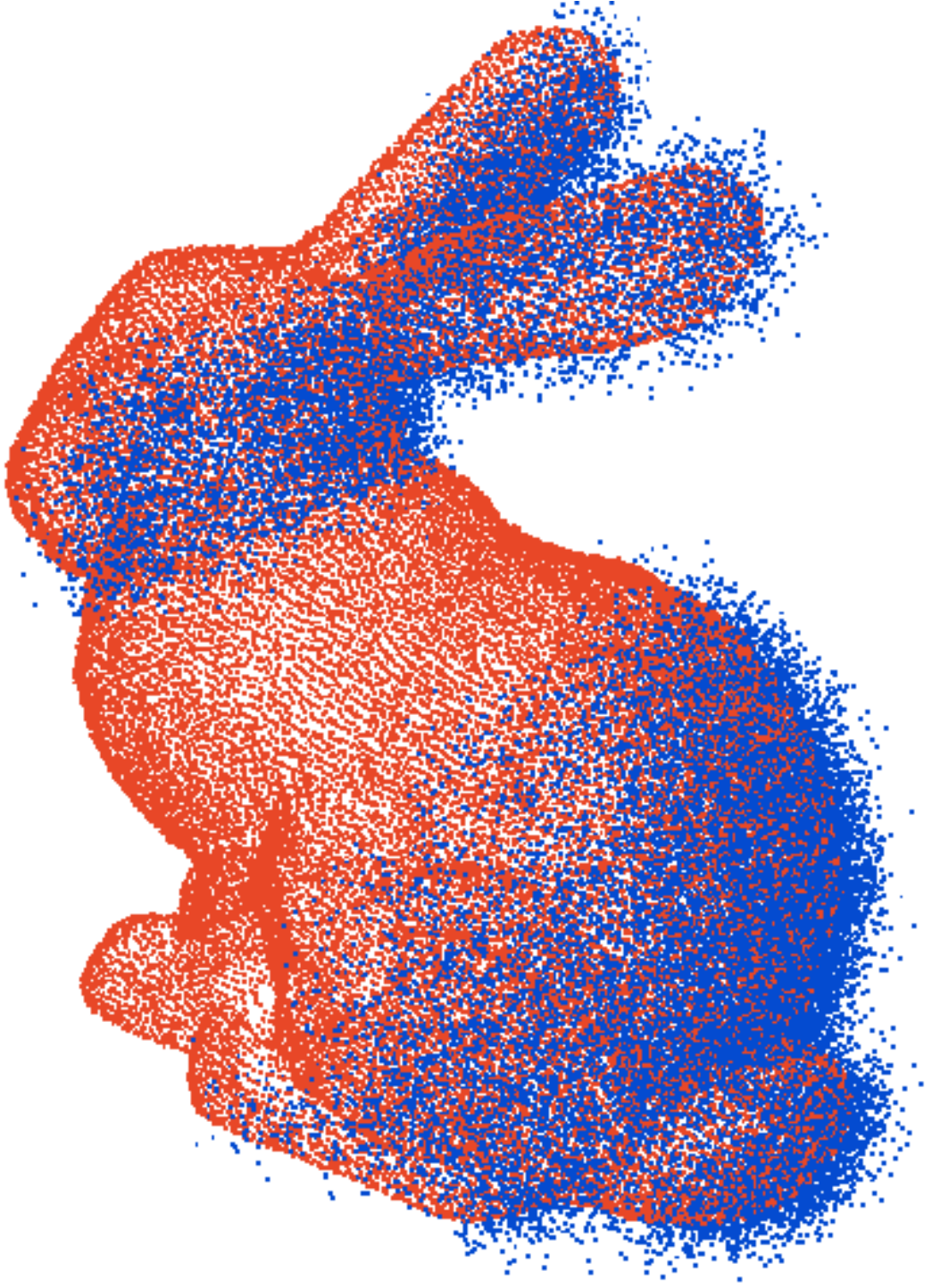}}~~
\subfigure{
\includegraphics[width=0.24\textwidth]{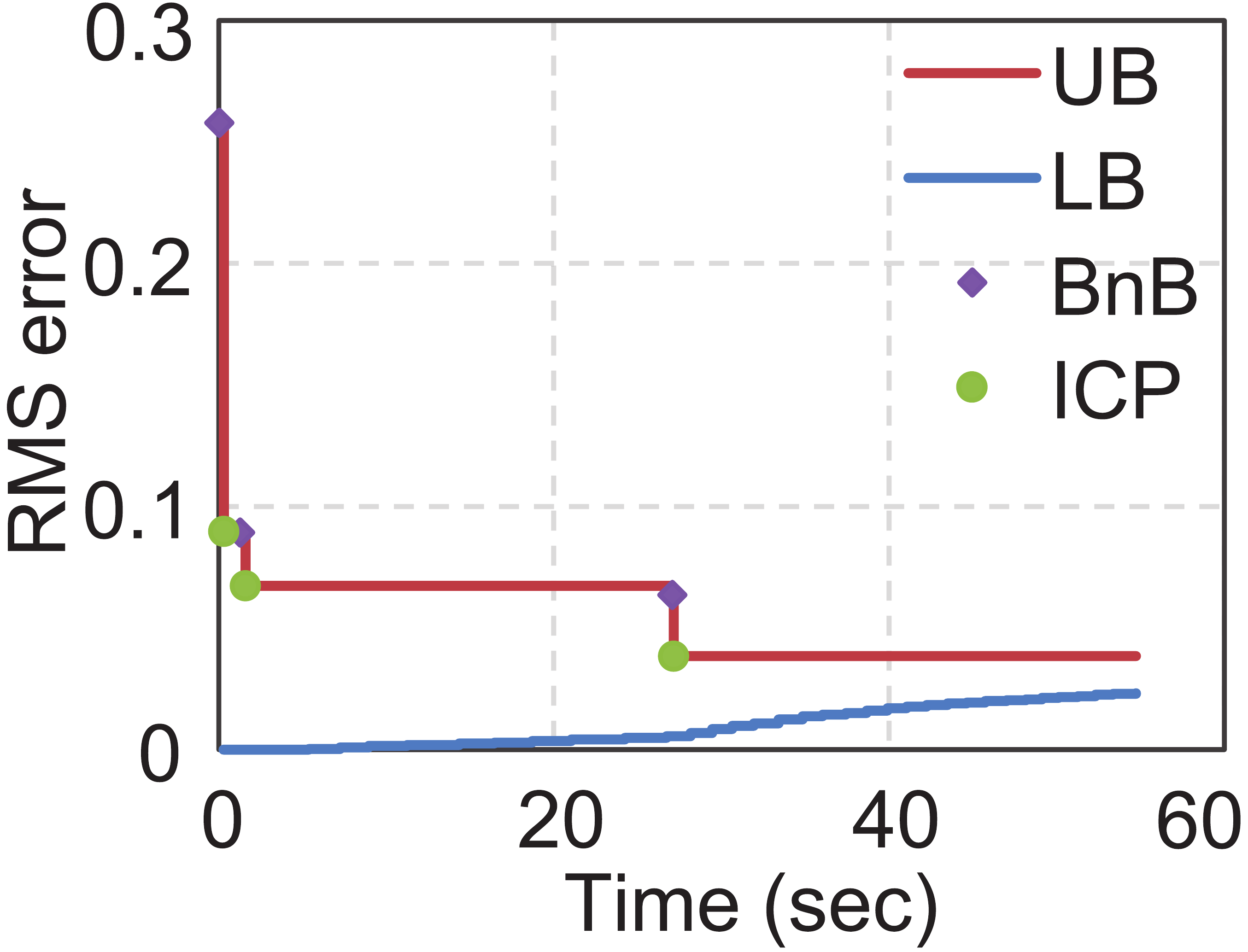}}
\vspace{-5pt}
\caption{Registration with high optimal error. \textbf{Left:} Gaussian noise was added to the data point-set to increase the RMS error. \textbf{Right:} the global minimum was found at about 25s with a DT; the remainder of the time was devoted solely to increasing the lower bound.
\label{fig:time_error_eg}}
\end{center}
\vspace{-7pt}
\end{figure}

\vspace{0.05in}
\noindent\textbf{Effect of Optimal Error.} We also tested the running time \emph{w.r.t.} the optimal registration error. To increase the error, Gaussian noise was added to the data point-set \emph{only}. As shown in Fig.~\ref{fig:time}, the running time remained almost constant when the RMS error was less than 0.03. This is because the gap between the global lower bound and the optimal error was less than $\epsilon$. Therefore, the running time depended primarily on when the global minimum was found, that is, the termination depended on the \emph{decrease of the upper bound}. However, it takes longer to converge if the final RMS error is higher. Figure~\ref{fig:time_error_eg} shows the bounds evolution for bunny when the RMS error was increased to $\sim\!0.04$. As can be seen, the global minimum was found at about 25s, with the remainder of the time devoted to \emph{increasing the lower bound}.

\subsection{Registration with Partial Overlap}\label{sec:outlierexp}

\begin{table}[!t]
    \centering
    \caption{Running time (in seconds) of Go-ICP with DTs for the registration of the partially overlapping point-sets in Fig.~\ref{fig:partialoverlap}. 100 random relative poses were tested for each point-set pair and 1\,000 data points were used. $\rho$ is the trimming percentage.}.
    \label{tab:time_partial}
    \begin{tabular}{|l|c|c|c|c|}
        \hline
         & \multicolumn{2}{|c|}{$A$$\rightarrow$$B$} &  \multicolumn{2}{|c|}{$B$$\rightarrow$$A$} \\
        \cline{2-5}
         & $\rho$  & \!\!mean/max time\!\! & $\rho$  & \!\!mean/max time\!\! \\
        \hline
        Bunny & 10\% & 0.81 / 10.7 & 10\% & 0.49 / 7.25 \\
        \hline
        Dragon  & 20\% & 2.99 / 43.5 &  40\% & 8.72 / 72.4 \\
        \hline
        Buddha  & 10\% & 0.71 / 11.3 & 10\% & 0.60 / 14.8 \\
        \hline
        Chef  & 20\% & 0.45 / 4.47 & 30\% & 0.52 / 3.79 \\
        \hline
        Dinosaur & 10\% & 2.03 / 23.5 & 10\% & 1.65 / 26.1 \\
        \hline
        Owl & 40\% & 12.5 / 87.5 & 40\% & 13.4 / 75.0 \\
        \hline
        Denture & 30\% & 6.74 / 74.7 & 30\% & 4.24 / 68.1 \\
        \hline
        Room & 30\% & 9.82 / 73.3 & 30\% & 18.4 /107.3 \\
        \hline
        Bowl & 20\% & 3.19 / 20.3 & 30\% & 3.52 / 25.3 \\
        \hline
        Loom & 30\% & 8.64 / 67.2 & 20\% & 5.96 / 44.6 \\
        \hline
    \end{tabular}
\vspace{-0pt}
\end{table}

In this section, we tested the proposed method on partially overlapping point-sets. The data points in regions that are not overlapped by the other model point-set should be treated as outliers, as their correspondences are missing. Trimming was employed to deal with outliers as described in Sec.~\ref{sec:outlier};

We used 10 point-set pairs shown in Fig.~\ref{fig:partialoverlap} to test Go-ICP with trimming. These point-sets were generated by different scanners and with different noise levels. The bunny, dragon and buddha models are from the Standford 3D dataset. The chef and dinosaur models are from \cite{mian2006three}. The denture was generated with a structured light 3D scanner\footnote{http://www.david-3d.com/en/support/downloads}. The owl status is from \cite{bouaziz2013sparse} and the room scans are from \cite{shotton2013scene}. The bowl and loom point-sets were collected by us with a Kinect. The overlapping ratio of the point-set pairs are between $50\%\!\sim\!95\%$.

\begin{figure*}[!t]
\begin{center}
\includegraphics[width=1.0\textwidth]{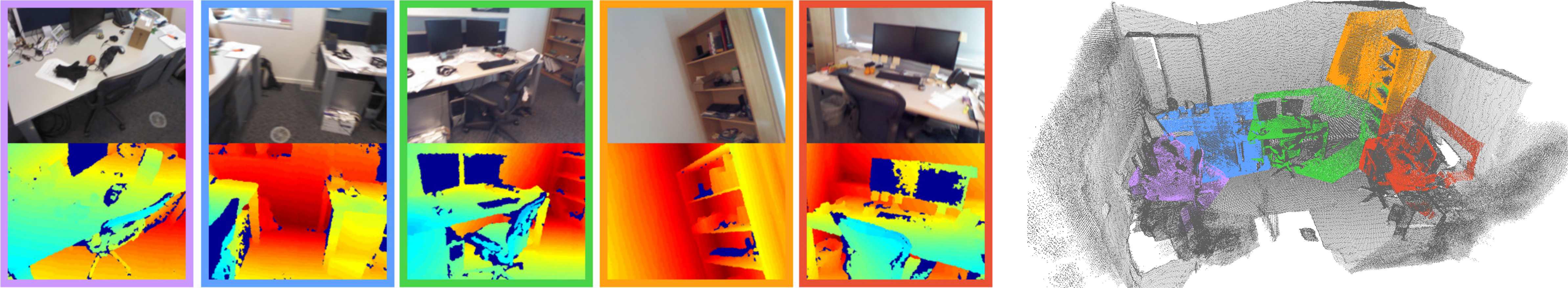}
\caption{Camera localization experiment. \textbf{Left}: 5 (out of 100) color and depth image pairs of the scene. (The color images were not used) \textbf{Right}: Corresponding registration results. Note that the scene contains many similar structures, and the depth images only cover small portions of the scene, which make the 3D
registration tasks very challenging.
\label{fig:localization}}
\end{center}
\vspace{-5pt}
\end{figure*}

For each of the 10 point-set pairs, we generated 100 random relative poses, and registered the two point-sets to each other. This lead to 2\,000 registration tasks. The translation domain to explore for Go-ICP was set to be $[-\pi,\pi]^3\times[-0.5,0.5]^3$. We chose the trimming percentages $\rho$ as in Table~\ref{tab:time_partial}, sampled $N\!=\!1000$ data points for each registration, and set all the convergence thresholds to $\epsilon\!=\!0.001\times K$ where $K=(1\!-\!\rho)\!\times\!N$. Our method correctly registered the point-sets in all these tasks. All the rotation errors were less than $5$ degrees and translation errors were less than $0.05$ compared to the manually-set ground truths. The running times using DTs are presented in Table~\ref{tab:time_partial}. In general, it takes the method a longer time compared to the outlier-free case due to 1) the emergence of additional local minima induced by the outliers and 2) the time-consuming trimming operations.

\vspace{0.06in}
\noindent\textbf{Choosing trimming percentages.} In these experiments, each parameter $\rho$ was chosen by visually observing the two point-sets and roughly guessing their non-overlapping ratios. The results were not very sensitive to $\rho$ (\eg setting $\rho$ as $5\%,10\%$ and $20\%$ all led to a successfully registration for bunny). If no rough guess is available, one can gradually increase $\rho$ until a measure such as the inlier number or RMS error attains a set value, or apply the automatic overlap estimation proposed in [62]. We also plan to test other outlier handling strategies (cf. Sec.~\ref{sec:outlier}) in future.

\subsection{More Applications}\label{sec:exp_application}
In this section, we present several additional scenarios where Go-ICP can be applied to achieve global optimality. Future efforts can be taken to extend the method and build complete real-world systems. In the following experiments, the transformation domain for exploration was set to be $[-\pi,\pi]^3\times[-1,1]^3$.

\vspace{0.06in}
\noindent\textbf{3D Object Localization.}
The proposed method is useful for model-based 3D object detection, localization and pose estimation from relatively large scenes. To experimentally verify this, we tested our method on one sequence of the camera localization dataset \cite{shotton2013scene}. Figure~\ref{fig:localization} shows a sample color and depth image pair, and a 3D model of the office scene. Our goal was to estimate the camera poses by registering the point clouds of the depth images onto the 3D scene model. We evenly sampled the sequence taken by a smoothly moving camera to 100 depth images. Each depth image was then sampled to $400\sim 600$ points. We set our method to seek a solution with the registration error smaller than $0.0001\!\times\!N$, and the method registered the 100 point-sets with the mean/longest running time of 32s/178s using a DT. The rotation errors and translation errors were all below 5 degrees and 10cm. Figure~\ref{fig:localization} shows 5 typical registration results.

\begin{figure}[!t]
\begin{center}
\includegraphics[width=0.49\textwidth]{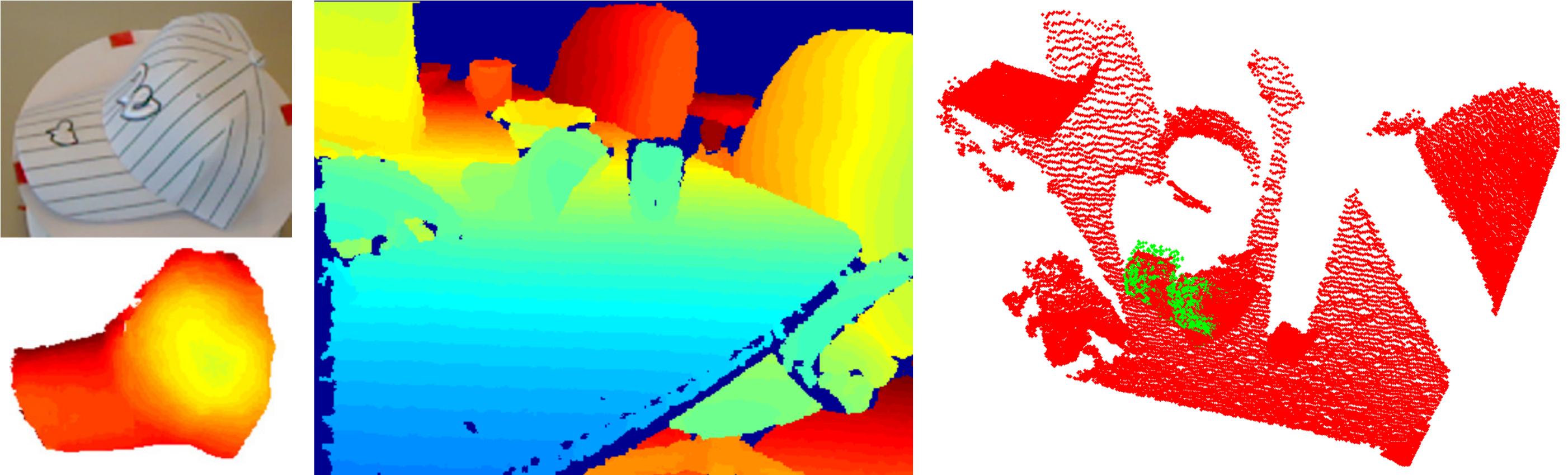}
\caption{3D object localization experiment. \textbf{Left}: a labelled object and its depth image to generate the data point-set. \textbf{Middle}: a scene depth image to generate the model point-set. \textbf{Right}: the registration result.
\label{fig:cap}}
\end{center}
\vspace{-9pt}
\end{figure}

We then used the RGB-D Object Dataset \cite{lai2011large}, with the goal of registering the points of a baseball cap to a point cloud of the scene, as shown in Fig.~\ref{fig:cap}. We sampled $N\!=\!100$ points from the cap model, and set the trimming percentage and threshold to be $\rho\!=\!10\%$ and $\epsilon\!=\!0.00003\!\times\!K$ respectively. Go-ICP successfully localized the cap in 42 seconds with a DT.

\begin{figure}[!t]
\begin{center}
\subfigure{
\includegraphics[width=0.156\textwidth]{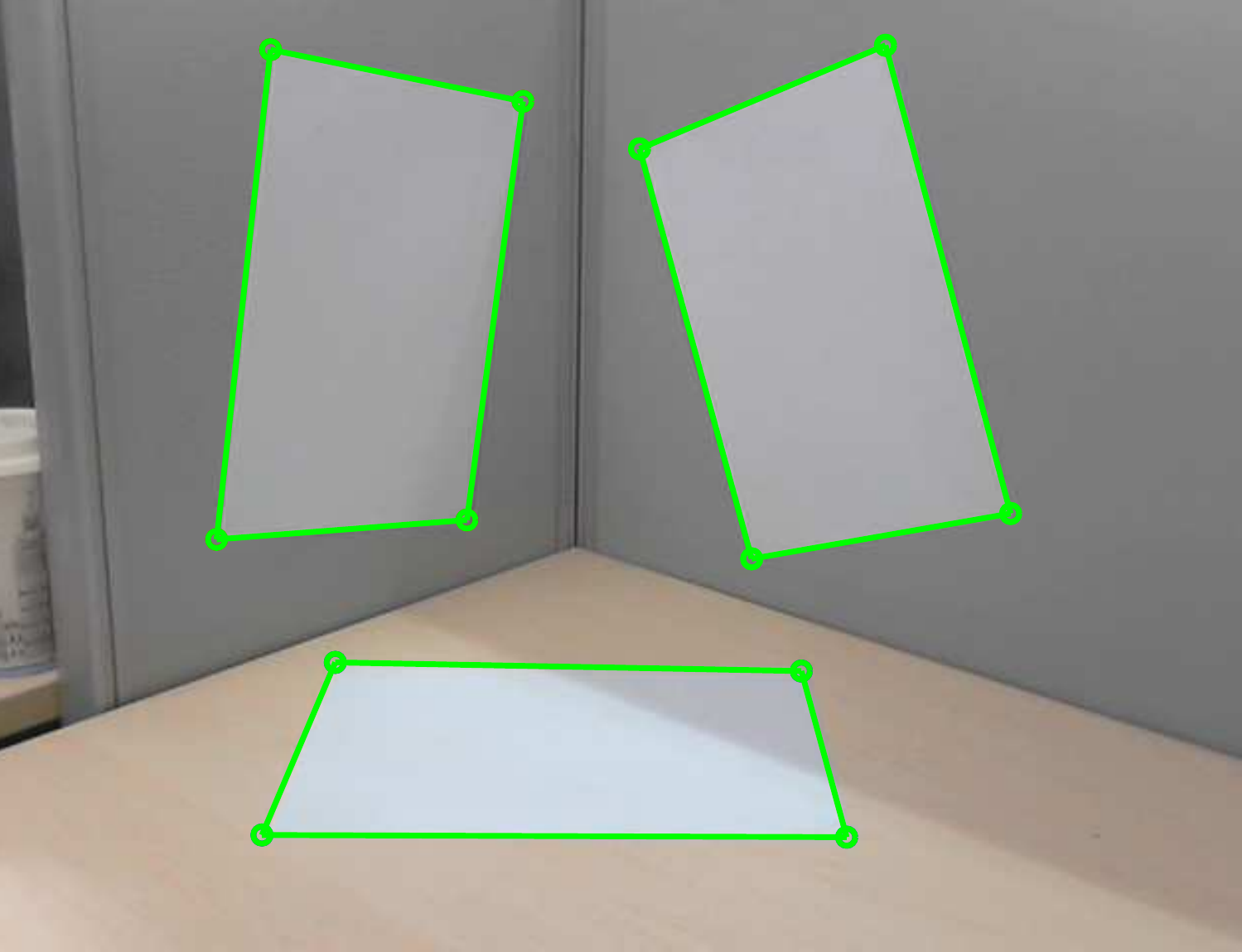}}\!\!
\subfigure{
\includegraphics[width=0.155\textwidth]{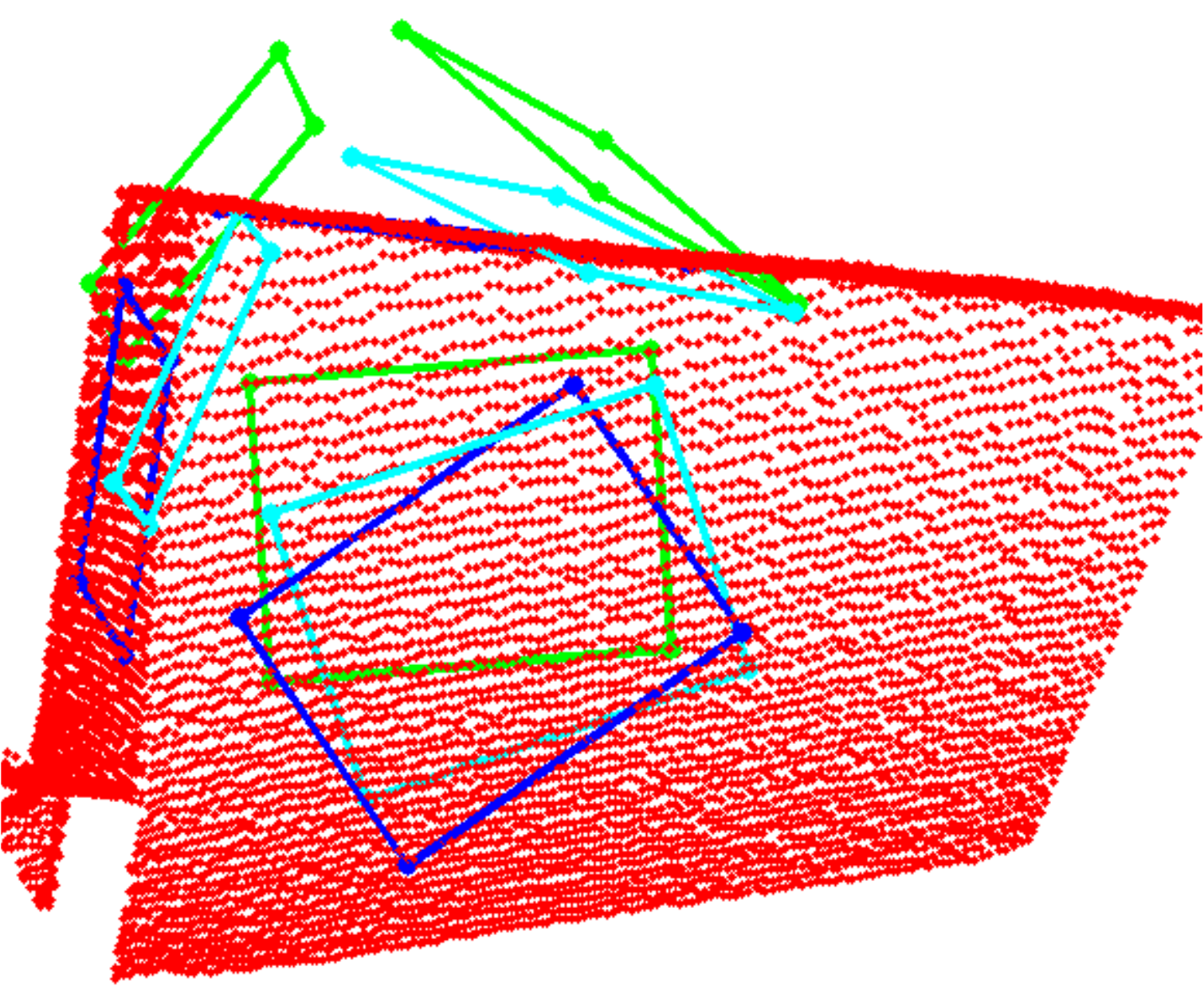}}\!\!
\subfigure{
\includegraphics[width=0.155\textwidth]{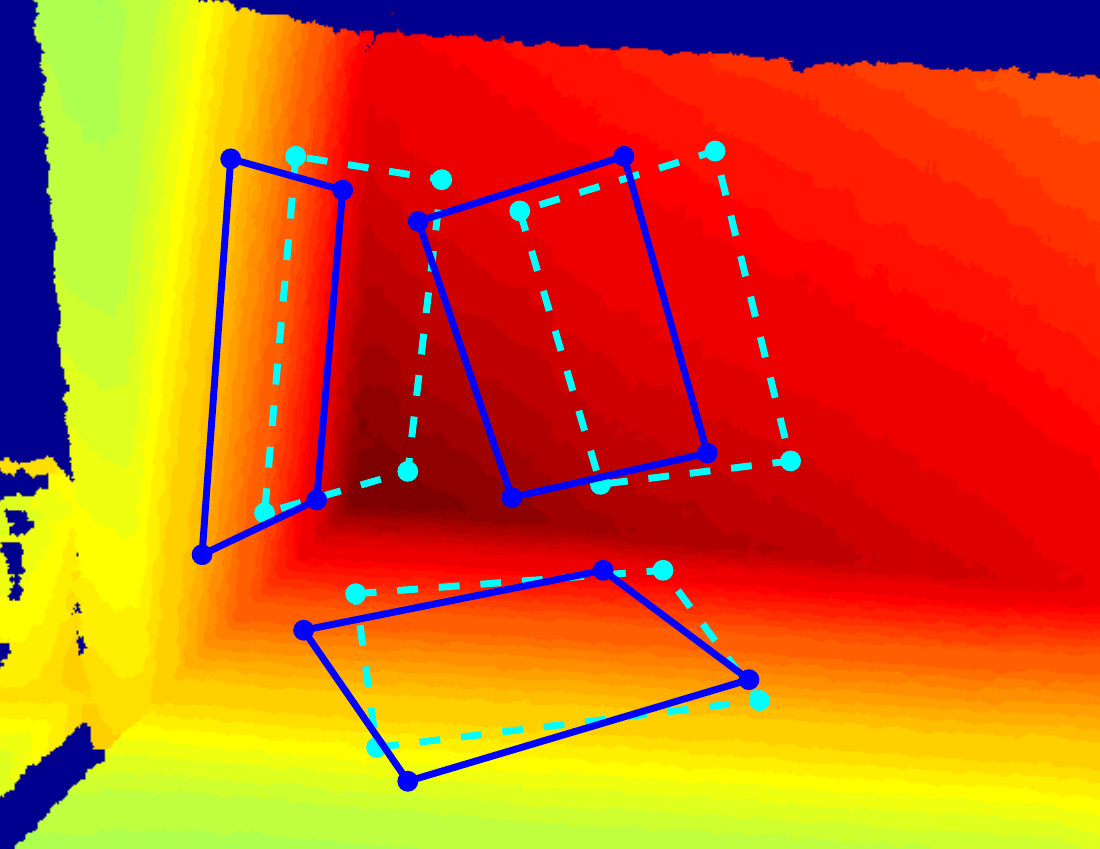}}
\caption{RGB-D extrinsic calibration experiment. \textbf{Left}: the color image with extracted line segments for single view 3D reconstruction. \textbf{Middle}: the initial 3D registration (in green), the result of ICP (in cyan) and the result of Go-ICP (in blue) (the lines are for visualization purposes only). \textbf{Right}: the depth image with a projection of the registered 3D points from ICP (in cyan) and Go-ICP (in blue).
\label{fig:calibration}}
\end{center}
\vspace{-9pt}
\end{figure}

\vspace{0.06in}
\noindent\textbf{Camera Extrinsic Calibration.}
In the work of Yang \etal~\cite{yang2013single}, the sparse point-set from a color camera, obtained by single view 3D reconstruction, was registered onto the dense point-set from a depth camera to obtain the camera relative pose. Figure~\ref{fig:calibration} shows an example where 12 points are reconstructed. We found that ICP often failed to find the correct registration when the pose difference between the cameras was reasonably large. To the best of our knowledge, few methods can perform such \emph{sparse-to-dense registration} reliably without human intervention, due to the difficulty of building putative correspondences.
Setting $\epsilon$ to be $0.00001\!\times\!N$, Go-ICP with a DT found the optimal solution in less than 1s. Note that the surfaces are not exactly perpendicular to each other.

\section{Conclusion}\label{sec:conclusion}
We have introduced a globally optimal solution to Euclidean registration in 3D, under the $L_2$-norm closest-point error metric originally defined in ICP. The method is based on the Branch-and-Bound (BnB) algorithm, thus global optimality is guaranteed regardless of the initialization. The key innovation is the derivation of registration error bounds based on the $SE(3)$ geometry.

The proposed Go-ICP algorithm is especially useful when an exactly optimal solution is highly desired or when a good initialization is not reliably available. For practical scenarios where real-time performance is not critical, the algorithm can be readily applied or used as an optimality benchmark.

\vspace{-5pt}
\ifCLASSOPTIONcompsoc
  \subsection*{Acknowledgments}
\else
  \subsection*{Acknowledgment}
\fi
{
This work was supported in part by the Natural Science Foundation of China (NSFC) under Grant No. 61375044, and ARC grants DP120103896, CE140100016 ARC Centre of Excellence for Robotic Vision. J. Yang was funded by Chinese Scholarship Council (CSC) from Sep 2013 to Aug 2015.
}

\ifCLASSOPTIONcaptionsoff
  \newpage
\fi

{
\bibliographystyle{IEEEtran}
\bibliography{Paper}
}


\vspace{-20pt}
\begin{IEEEbiography}[{\includegraphics[width=1in,height=1.25in,clip,keepaspectratio]{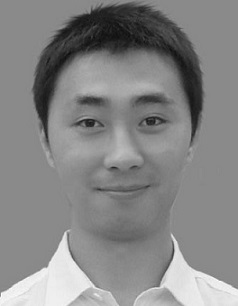}}]{Jiaolong Yang}
received a BS degree in Computer Science from the Beijing Institute of Technology (BIT) in 2010. He is now a dual PhD candidate at the Beijing Laboratory of Intelligent Information Technology, BIT, and the Research School of Engineering, Australian National University (ANU). He was awarded a Chinese Government Scholarship by the China Scholarship Council to study at ANU from September 2013 to September 2015. His current research interests include 3D and geometric computer vision.
\vspace{-25pt}
\end{IEEEbiography}
\begin{IEEEbiography}[{\includegraphics[width=1in,height=1.25in,clip,keepaspectratio]{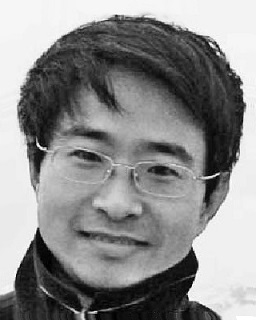}}]{Hongdong Li}
is faculty member with Research School of Engineering, the Australian National University (ANU) and National ICT Australia.  He joined the ANU since 2004 after graduated from Zhejiang University in China.  His current research interests include vision geometry,  3D computer vision and optimization.  He serves various programme committees for major computer vision conferences, including being an Area Chair for ICCV'13, ECCV'14, CVPR'15 and ICCV'15.  He is a regular reviewer for TPAMI and IJCV. He was a winner for the CVPR'12 Best Paper Award, ICPR'10 Best Student Paper Award, and ICIP'14 Best Student Paper Award.
\vspace{-25pt}
\end{IEEEbiography}
\begin{IEEEbiography}[{\includegraphics[width=1in,height=1.25in,clip,keepaspectratio]{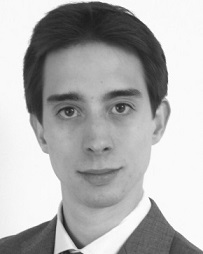}}]{Dylan Campbell}
received a BE degree in Mechatronic Engineering from the University of New South Wales in Sydney, Australia in 2012. He is currently a Computer Vision PhD student at the Australian National University and National ICT Australia. His research interests include place recognition, global localisation, mobile robotics, 3D reconstruction, multi-sensor perception and global optimization.
\vspace{-25pt}
\end{IEEEbiography}
\begin{IEEEbiography}[{\includegraphics[width=1in,height=1.25in,clip,keepaspectratio]{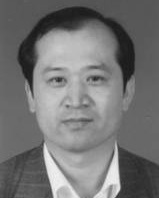}}]{Yunde Jia} received the M.S. and Ph.D. degrees in mechatronics from the Beijing Institute of Technology (BIT), Beijing, China, in 1986 and 2000, respectively. He is currently a Professor of computer science with BIT, and serves as the Director of the Beijing Laboratory of Intelligent Information Technology, School of Computer Science. He has previously served as the Executive Dean of the School of Computer Science, BIT, from 2005 to 2008. He was a Visiting Scientist at Carnegie Mellon University, Pittsburgh, PA, USA, from 1995 to 1997, and a Visiting Fellow at the Australian National University, Acton, Australia, in 2011. His current research
interests include computer vision, media computing, and intelligent systems.
\end{IEEEbiography}




\end{document}